\documentclass[runningheads]{llncs}
\usepackage{graphicx}
\usepackage{comment}
\usepackage{amsmath,amssymb} 
\usepackage{color}
\usepackage{tabularx}
\usepackage{latexsym}
\usepackage{makecell}
\usepackage[width=122mm,left=12mm,paperwidth=146mm,height=193mm,top=12mm,paperheight=217mm]{geometry}

\usepackage{comment}
\usepackage{multirow}
\usepackage{multicol}
\usepackage{amssymb}
\usepackage{pifont}
\newcommand{\cmark}{\ding{51}}%

\begin{document}

\pagestyle{headings}
\mainmatter

\title{A Closer Look at Local Aggregation Operators in Point Cloud Analysis} 

\titlerunning{A Closer Look at 3D Local Aggregation Operators}
%
\author{Ze Liu\inst{1,2}\thanks{Equal contribution. $^\dag$This work is done when Ze Liu is an intern at MSRA.}$^\dag$ \and
Han Hu\inst{2\star} \and
Yue Cao\inst{2} \and Zheng Zhang\inst{2} \and Xin Tong\inst{2}}
\authorrunning{Liu et al.}
%
\institute{University of Science and Technology of China\\
\email{liuze@mail.ustc.edu.cn} \and
Microsoft Research Asia\\
\email{\{hanhu,yuecao,zhez,xtong\}@microsoft.com}}

%
\maketitle

\begin{abstract}
Recent advances of network architecture for point cloud processing are mainly driven by new designs of local aggregation operators. However, the impact of these operators to network performance is not carefully investigated due to different overall network architecture and implementation details in each solution. Meanwhile, most of operators are only applied in shallow architectures. In this paper, we revisit the representative local aggregation operators and study their performance using the same deep residual architecture. Our investigation reveals that despite the different designs of these operators, all of these operators make surprisingly similar contributions to the network performance under the same network input and feature numbers and result in the state-of-the-art accuracy on standard benchmarks. This finding stimulate us to rethink the necessity of sophisticated design of local aggregation operator for point cloud processing. To this end, we propose a simple local aggregation operator without learnable weights, named Position Pooling (PosPool), which performs similarly or slightly better than existing sophisticated operators. In particular, a simple deep residual network with PosPool layers achieves outstanding performance on all benchmarks, which outperforms the previous state-of-the methods on the challenging PartNet datasets by a large margin (7.4 mIoU). The code is publicly available at \url{https://github.com/zeliu98/CloserLook3D}.

\end{abstract}

\section{Introduction}

With the rise of 3D scanning devices and technologies, 3D point cloud becomes a popular input for many machine vision tasks, such as autonomous driving, robot navigation, shape matching and recognition, etc. Different from images and videos that are defined on regular grids, the point cloud locates at a set of irregular positions in 3D space, which makes the powerful convolutional neural networks (CNN) and other deep neural networks designed for regular data hard to be applied. Early studies transform the irregular point set into a regular grid by either voxelization or multi-view 2D projections such that the regular CNN can be adopted. However, the conversion process always results in extra computational and memory costs and the risk of information loss.

Recent methods in point cloud processing develop networks that can directly model the unordered and non-grid 3D point data. These architectures designed for point cloud are composed by two kinds of layers: the point-wise transformation layers and local aggregation layers. While the point-wise transformation layer is applied on features at each point, the local aggregation layer plays a similar role for points  as the convolution layer does for image pixels. Specifically, it takes features and relative positions of neighborhood points to a center point as input, and outputs the transformed feature for the center point. To achieve better performance in different point cloud processing tasks, a key task of point cloud network design is to develop effective local aggregation operators.

Existing local aggregation operators can be roughly categorized into three groups according to the way that they combine the relative positions and point features: point-wise multi-layer perceptions (MLP) based~\cite{qi2017pointnet++,wang2019dynamic,li2019can,komarichev2019cnn}, pseudo grid feature based~\cite{hua2018pointwise,mao2019interpolated,zhang2019shellnet,lan2019modeling,tatarchenko2018tangent,thomas2019kpconv} and adaptive weight based~\cite{wang2018paramconv,groh2018flex,liu2019rscnn,wu2019pointconv,wang2019graph,li2018pointcnn}. The point-wise MLP based methods treat a point feature and its corresponding relative position equally by concatenation. All the concatenated features at neighborhood are then abstracted by a small PointNet~\cite{qi2017pointnet} (multiple point-wise transformation layers followed by a MAX pooling layer) to produce the output feature for the center point. The pseudo grid feature based methods first generate pseudo features on pre-defined grid locations, and then learn the parametrized weights on these grid locations like regular convolution layer does. The adaptive weight based methods aggregate neighbor features by weighted average with the weights adaptively determined by relative position of each neighbor.

Despite the large efforts for aggregation layer design and performance improvements of the resulting network in various point cloud processing tasks, the contributions of the aggreation operator to the network performance have never been carefully investigated and fairly compared. This is mainly due to the different network architectures used in each work, such as the network depth, width, basic building blocks, whether to use skip connection, as well as different implementation of each approach, such as point sampling method, neighborhood computation, and so on. Meanwhile, most of existing aggregation layers are applied in shallow networks, it is unclear whether these designs are still effective as the network depth increases.

In this paper, we present common experimental settings for studying these operators, selecting a deep residual architecture as the base networks, as well as same implementation details regarding point sampling, local neighborhood selection and etc. We also adopt three widely used datasets, ModelNet40~\cite{wu20153d}, S3DIS~\cite{2017arXiv170201105A} and PartNet~\cite{mo2019partnet} for evaluation, which account for different tasks, scenarios and data scales. Using these common experimental settings, we revisit the performance of each representative operator and make fair comparison between them. We find appropriate settings for some operators under this deep residual architecture are different from that of using shallower and non-residual networks. We also surprisingly find that different representative methods perform similarly well under the same representation capacity on these datasets, if appropriate settings are adopted for each method, although these methods may be invented by different motivations and formulations, in different years.

These findings also encourage us to rethink the role of local aggregation layers in point cloud modeling: \emph{do we really need sophisticated/heavy local aggregation computation?} We answer this question by proposing an extremely simple local aggregation operator with no learnable weights: combining a neighbor point feature and its 3-d relative coordinates by element-wise multiplication, followed with an AVG pool layer to abstract information from neighborhood. We name this new operator as position pooling (PosPool), which shows no less or even better accuracy than other highly tuned sophisticated operators on all the three datasets. These results indicate that we may not need sophisticated/heavy operators for local aggregation computation. We also harness a strong baseline for point cloud analysis by a simple deep residual architecture and the proposed position pooling layers, which achieves 53.8 part category mIoU accuracy on the challenging PartNet datasets, significantly outperforming the previous best method by 7.4 mIoU.

The contributions of this paper are summarized as
\begin{itemize}
    \item \textbf{A common testbed} to fairly evaluate different local aggregation operators.
    \item \textbf{New findings of aggregation operators.} Specifically, \emph{different operators perform similarly well and all of them can achieve the state-of-the-art accuracy}, if appropriate settings are adopted for each operator. Also, \emph{appropriate settings in deep residual networks are different from those in shallower networks}. We hope these findings could shed new light on network design.
    \item \textbf{A new local aggregation operator (PosPool) with no learnable weights} that performs as effective as existing operators. Combined with a deep residual network, this simple operator achieve state-of-the-art performance on 3 representative benchmarks and outperforms the previous best method by a large margin of 7.4 mIoU on the challenging PartNet datasets.
\end{itemize}

\section{Related Works}

\vspace{0.3em} \noindent \textbf{Projection based Methods} project the irregular point cloud onto a regular sampling grid and then apply 2D or 3D CNN over regularly-sampled data for various vision tasks. View-based methods project a 3D point cloud to a set of 2D views from various angles. Then these view images could be processed by 2D CNNs \cite{feng2018gvcnn,guo2016multiview3d,qi2016volumetric,su2015multi}. Voxel-based methods project the 3D points to regular 3D grid, and then standard 3D CNN could be applied \cite{gadelha2018multiresolution,maturana2015voxnet,wu20153d}. Recently, adaptive voxel-based representations such as K-d trees \cite{klokov2017kdnet} or octrees \cite{riegler2017octnet,tatarchenko2017octree,wang2017ocnn} have been proposed for reducing the memory and computational cost of 3D CNN. The view-based and voxel-based representations are also combined ~\cite{qi2016volumetric} for point cloud analysis. All these methods require preprocessing to convert the input point cloud and may lose the geometry information.

\vspace{0.3em} \noindent \textbf{Global Aggregation Methods} process the 3D point cloud via point-wise $1\times 1$ transformation (fully connected) layers followed by a global pooling layer to aggregate information globally from all points~\cite{qi2017pointnet}. These methods are the first to directly process the irregular point data. They have no restriction on point number, order and regularity of neighborhoods, and obtain fairly well accuracy on several point cloud analysis tasks. However, the lack of local relationship modeling components hinders the better performance on these tasks.

\vspace{0.3em} \noindent \textbf{Local Aggregation Methods} Recent point cloud architectures are usually composed by $1\times 1$ point-wise transformation layers and local aggregation operators. Different methods are mainly differentiated by their local aggregation layers, which usually adopt the neighboring point features and their relative coordinates as input, and output a transformed center point feature. According to the way they combine point features and relative coordinates, these methods can be roughly categorized into three groups: point-wise MLP based~\cite{qi2017pointnet++,li2019can,komarichev2019cnn}, pseudo grid feature based~\cite{hua2018pointwise,mao2019interpolated,zhang2019shellnet,lan2019modeling,tatarchenko2018tangent,thomas2019kpconv}, and adaptive weight based~\cite{wang2018paramconv,groh2018flex,liu2019rscnn,wu2019pointconv,li2018pointcnn}, as will be detailed in Section~\ref{sec.review}. There are also some works use additional edge features (relative relationship between point features) as input~\cite{li2019can,wang2019graph,wang2019dynamic}, also commonly referred to as graph based methods.

While we have witnessed significant accuracy improvements on benchmarks by new local aggregation operators year-by-year, the actual progress is a bit vague to the community as the comparisons are made on different grounds that the other architecture components and implementations may vary significantly. The effectiveness of designing components in some operators using deeper residual architectures is also unknown.

\section{Overview of Local Aggregation Operators}
\label{sec.review}

In this section, we present a general formulation for local aggregation operators as well as a categorization of them.

\vspace{0.3em} \noindent \textbf{General Formulation} In general, for each point $i$, a local aggregation layer first transforms a neighbor point $j$'s feature $\mathbf{f}_j\in \mathbb{R}^{d\times 1}$ and its relative location $\Delta \mathbf{p}_{ij} = \mathbf{p}_j - \mathbf{p}_i\in \mathbb{R}^{3\times 1}$ into a new feature by a function $G(\cdot, \cdot)$, and then aggregate all transformed neighborhood features to form point $i$'s output feature by a reduction function $R$ (typically using MAX, AVG or SUM), as
\begin{equation}
\label{eq:local_aggregation}
    \mathbf{g}_i = R\left(\left\{G(\Delta \mathbf{p}_{ij}, \mathbf{f}_j) | {j \in \mathcal{N}(i)}\right\}\right),
\end{equation}
where $\mathcal{N}(i)$ represents the neighborhood of point $i$. Alternatively, edge features $\{\mathbf{f}_i, \Delta \mathbf{f}_{ij} \}$ ($\Delta \mathbf{f}_{ij} = \mathbf{f}_j - \mathbf{f}_i$) can be used as input instead of $\Delta \mathbf{p}_{ij}$~\cite{wang2019dynamic}.

According to the family to which the transformation function $G(\cdot, \cdot)$ belongs, existing local aggregation operators can be roughly categorized into three types: point-wise MLP based, pseudo grid feature based, and adaptive weight based.

\vspace{0.3em} \noindent \textbf{Point-wise MLP based Methods} The pioneer work of point-wise MLP based method, PointNet++~\cite{qi2017pointnet++}, applies several point-wise transformation (fully connected) layers on a concatenation of relative position and point feature to achieve transformation:
\begin{equation}
\label{eq:pw_mlp}
G(\Delta \mathbf{p}_{ij}, \mathbf{f}_j) = \text{MLP}\left(\text{concat}(\Delta \mathbf{p}_{ij}, \mathbf{f}_j)\right).
\end{equation}
There are also variants by using an alternative edge feature $\{\mathbf{f}_i, \Delta \mathbf{f}_{ij}\}$ as input~\cite{wang2019dynamic,li2019can}, or by using a special neighborhood strategy~\cite{komarichev2019cnn}. The reduction function $R(\cdot)$ is usually set as MAX~\cite{qi2017pointnet++,wang2019dynamic,li2019can}.

The multiple point-wise layers after concatenation operation can approximate any continuous function about the relative coordinates and point feature~\cite{qi2017pointnet,qi2017pointnet++}. However, a drawback lies in its large computation complexity, considering the fact that the multiple fully connected (FC) layers are applied to all neighboring points when computing each point's output. Specifically, the FLOPs is $\mathcal{O}(\text{time})$ $=((2d + 3) +(h-2)d/2)\cdot d/2 \cdot n K$, for a point cloud with $n$ points, neighborhood size of $K$, FC layer number of $h$, and inter-mediate dimension of $d/2$, when $h \geq 2$. The space complexity is $\mathcal{O}(\text{space})=((2d + 3) +(h-2)d/2)\cdot d/2$. For $h = 1$, there exists efficient implementation by computation sharing (see Section~\ref{sec.benchmark_mlp}).

\vspace{0.3em} \noindent \textbf{Pseudo Grid Feature based Methods} The pseudo grid feature based methods generate pseudo features on several sampled regular grid points, such that regular convolution methods can be applied. A representative method is KPConv~\cite{thomas2019kpconv}, where equally distributed spherical grid points are sampled and the pseudo features on the $k^\text{th}$ grid point is computed as
\begin{equation}
\label{eq:pseudo_points}
\begin{small}
\textbf{f}_{i,k} = \sum\limits_{j \in \mathcal{N}(i)} \max(0,1-\frac{\|\Delta \mathbf{p}_{jk}\|_2}{\sigma}) \cdot \mathbf{f}_{j}.
\end{small}
\end{equation}
The index of each grid point $k$ will have strict mapping with the relative position to center point $\Delta \mathbf{p}_{ik}$. Hence, a (depth-wise) convolution operator with parametrized weights $\mathbf{w}_k \in \mathbb{R}^{d \times 1}$ defined on each grid point can be used to achieve feature transformation:
\begin{equation}
G(\Delta \mathbf{p}_{ik}, \mathbf{f}_{i,k}) = \mathbf{w}_k \odot \mathbf{f}_{i,k}.
\end{equation}

\begin{figure}[t]
    \centering
    \includegraphics[width=0.9\linewidth]{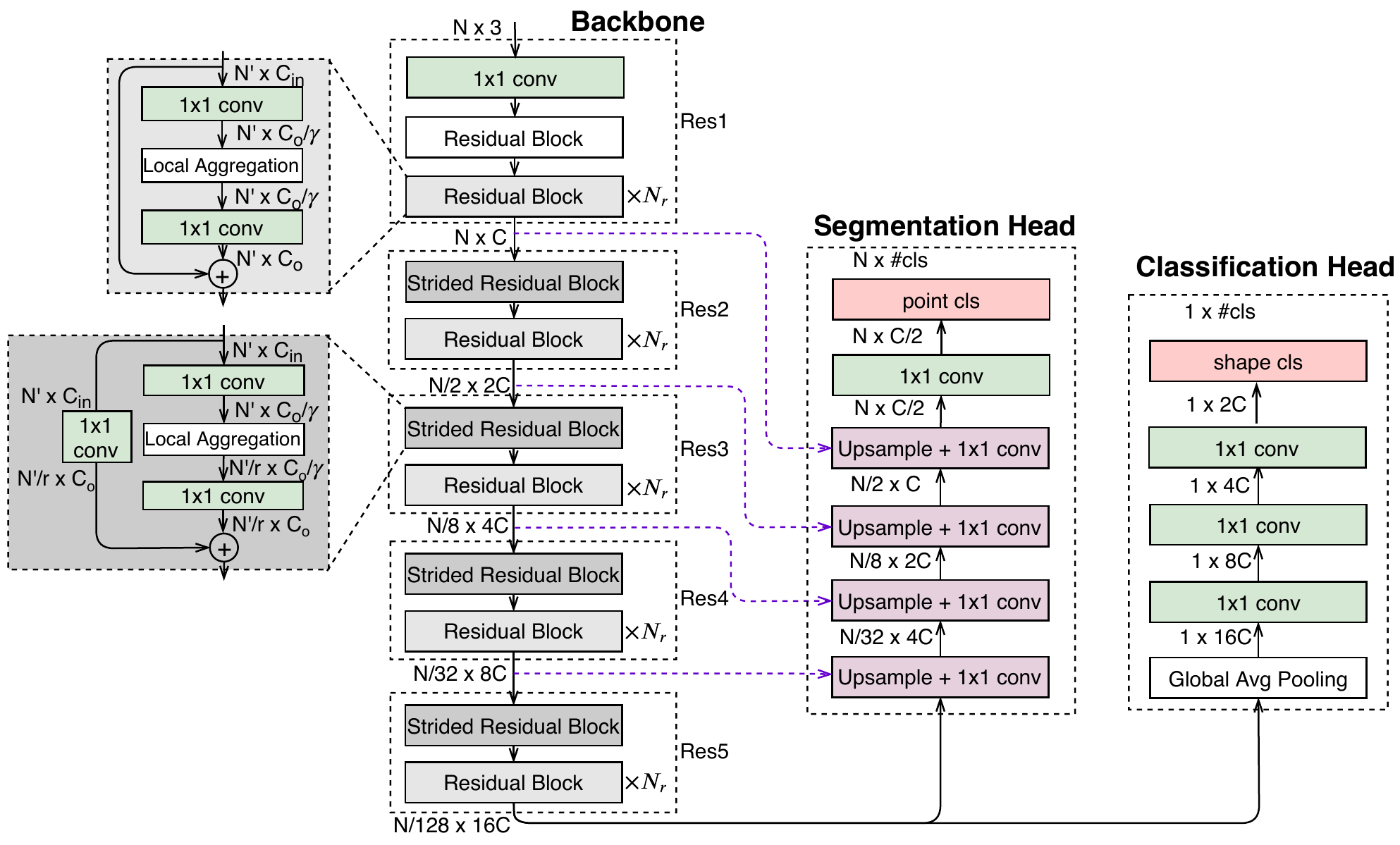}
    \vspace{-1.5em}
    \caption{A common deep residual architecture used to evaluate different local aggregation operators. In evaluation, we adjust the model complexity by changing architecture depth (or block repeating factor $N_r$), base width $C$ and bottleneck ratio $\gamma$. Note the point numbers drawn in this figure is an approximation to indicate the rough complexity but not an accurate number. Actually, the points on each stage are generated by a subsampling method~\cite{Thomas_2018} using a fixed grid size and the point number on different point cloud instances can vary}
    \label{fig:arch}
    \vspace{-1.5em}
\end{figure}

Different pseudo grid feature based methods mainly differ each other by the definition of grid points~\cite{hua2018pointwise,mao2019interpolated,zhang2019shellnet,lan2019modeling,tatarchenko2018tangent} or index order~\cite{li2018pointcnn}. When depth-wise convolution is used, the space and time complexity are $\mathcal{O}(\text{space})$ $= dM$ and $\mathcal{O}(\text{time}) = ndKM$, respectively, where $M$ is the number of grid points.

\vspace{0.3em} \noindent \textbf{Adaptive Weight based Methods} The adaptive weight based methods define convolution filters over arbitrary relative positions, and hence can compute aggregation weights on all neighbor points:
\begin{equation}
\label{eq:continuous_filter}
G(\Delta \mathbf{p}_{ij}, \mathbf{f}_j) = H\left(\Delta \mathbf{p}_{ij} \right) \odot \mathbf{f}_j,
\end{equation}
where $H$ is typically implemented by several point-wise transformation (fully connected) layers~\cite{wang2018paramconv,groh2018flex}; $\odot$ is an element-wise multiplication operator; $R$ is typically set as SUM.

Some methods adopt more position related variables~\cite{liu2019rscnn}, point density~\cite{wu2019pointconv}, or edge features~\cite{wang2019graph} as the input to compute adaptive weights. More sophisticated function other than fully connected (FC) layers are also used, for example, Taylor approximation~\cite{li2018pointcnn} and an additional SoftMax function to normalize aggregation weights over neighborhood~\cite{wang2019graph}.

The space and time complexity of this method are $\mathcal{O}(\text{space}) = ((h-2)d/2 + d + 3) \cdot d/2$ and $\mathcal{O}(\text{time}) = ((h-2)d/2 + d + 5) \cdot d/2 \cdot nK$, respectively, when an inter-mediate dimesion of $d/2$ is used and the number of FC layers $h \geq 2$. When $h=1$, the space and computation complexity is much smaller, as $\mathcal{O}(\text{space}) = 3d$ and $\mathcal{O}(\text{time}) = 5dnK$, respectively.

Please see Appendix A6 for detailed analysis of the space and time complexity for the above 3 operators.

\section{Benchmarking Local Aggregation Operators in Common Deep Architecture}

While most local aggregation operators described in Section~\ref{sec.review} are reported using specific shallow architectures, it is unknown whether their designing components perform also sweet using a deep residual architecture. In addition, these operators usually use different backbone architectures and different implementation details, making a fair comparison between them difficult.

In this section, we first present a deep residual architecture, as well as implementation details regarding point sampling and neighborhood selection. Then we evaluate the designing components of representative operators using common architectures, implementation details and benchmarks. The appropriate settings within each method type using the common deep residual architectures are suggested and discussed.

\subsection{Common Experimental Settings}

\vspace{0.3em} \noindent \textbf{Architecture} To investigate different local aggregation operators on a same, deep and modern ground, we select a 5-stage deep residual network, similar to the standard ResNet model~\cite{he2015deep} in image analysis. Residual architectures have been widely adopted in different fields to facilitate the training of deep networks~\cite{he2015deep,vaswani2017attention}. However, in the point cloud field, until recently, there are some works~\cite{thomas2019kpconv,li2019can} starting to use deep residual architectures, probably because the unnecessary use of deep networks on several small scale benchmarks. Nevertheless, our investigation shows that on larger scale and more challenging datasets such as PartNet~\cite{mo2019partnet}, deep residual architectures can bring significantly better performance, for example, with either local aggregation operator type described in Section~\ref{sec.review}, the deep residual architectures can surpass previous best methods by more than 3 mIoU. On smaller scale datasets such as ModelNet40, they also seldom hurt the performance. The deep residual architecture would be a reasonable choice for practitioners working on point cloud analysis.

\begin{small}
\begin{table}[t]
    \centering
    \caption{The performance of baseline operators, sweet spots of point-wise MLP based, pseudo grid feature based and adaptive weight based operators, and the proposed PosPool operators on three benchmark datasets. Baseline$^*$ denotes Eq.~(\ref{eq.no_local}) and baseline$^\dag$ (AVG/MAX) denotes Eq.~(\ref{eq.max_avg_pool}) AVG/MAX, respectively. PosPool and PosPool* denote the operators in Eq.~(\ref{eq:pos_pool}) and (\ref{eq:pos_pool_cosine_sine}), respectively. (S) after each method denotes a smaller configuration of this method ($N_r = 1$, $\gamma=2$ and $C=36$), which is about $16\times$ more efficient than the regular configuration (the other row) of $N_r = 1$, $\gamma=2$ and $C=144$. Previous best performing methods on three benchmarks in literature are shown in the first block of this table}
    \label{tab:baseline}
\addtolength{\tabcolsep}{.1pt}
\begin{tabular}{l|ccc|ccc|cccc}
\Xhline{1.0pt}
\multirow{2}{*}{method} &  \multicolumn{3}{c|}{ModelNet40} & \multicolumn{3}{c|}{S3DIS} & \multicolumn{4}{c}{\makecell{PartNet}} \\
\cline{2-11}
& acc & param & FLOP & mIoU & param & FLOP & val & test & param & FLOP \\
\Xhline{1.0pt}
DensePoint~\cite{liu2019densepoint} & 93.2 &0.7M & 0.7G& - & -&- & -&- &- &- \\
KPConv~\cite{thomas2019kpconv} & 92.9& 15.2M&1.7G & 65.7 &15.0M & 6.5G& -&- & -&- \\
PointCNN~\cite{li2018pointcnn} & 92.5& 0.6M& 25.3G& 65.4&4.4M &36.7G & - & 46.4 &4.4M &23.1G \\
\Xhline{1.0pt}
baseline$^*$ & 91.4 &19.4M &1.8G & 51.5& 18.4M& 7.2G& 42.5& 44.6 &18.5M & 6.7G\\
\hline
baseline$^\dag$ (AVG, S)  & 90.7& 1.2M& 0.1G& 50.3&  1.1M& 0.5G& 39.5&40.6 & 1.1M&0.4G\\
baseline$^\dag$ (AVG) & 91.4& 19.4M&1.8G & 51.0& 18.4M&7.2G & 44.2& 45.8&18.5M & 6.7G\\
\hline
baseline$^\dag$ (MAX, S)  &91.5 & 1.2M& 0.1G& 57.4&  1.1M&0.5G &39.8 &41.2 & 1.1M&0.4G\\
baseline$^\dag$ (MAX) & 91.8 & 19.4M& 1.8G& 58.4& 18.4M& 7.2G&  45.4& 47.4 & 18.5M& 6.7G\\
\Xhline{1.0pt}
point-wise MLP (S)  & 92.6&1.7M & 0.2G& 56.7&  1.6M& 0.8G& 45.3&47.0 & 1.6M&0.7G\\
point-wise MLP & 92.8& 26.5M&2.7G & 66.2& 25.5M& 9.8G& 48.1&51.5 & 25.6M& 9.1G\\
\hline
pseudo grid (S)  & 92.3& 1.2M& 0.3G& 64.3& 1.2M & 1.0G& 44.2&45.2 & 1.2M&0.9G\\
pseudo grid & 93.0& 19.5M&2.0G& 65.9& 18.5M& 9.3G&50.8 &53.0 & 18.5M& 8.5G\\
\hline
adapt weights (S) & 92.1& 1.2M& 0.2G& 61.9&  1.2M& 0.6G& 44.1& 46.1& 1.2M&0.5G\\
adapt weights & 93.0& 19.4M& 2.3G& 66.5& 18.4M& 7.8G& 50.1& 53.5& 18.5M& 7.2G\\
\Xhline{1.0pt}
PosPool (S)   & 92.5& 1.2M& 0.1G& 64.2&  1.1M& 0.5G& 44.6&47.2 &1.1M &0.5G\\
PosPool & 92.9& 19.4M& 1.8G& 66.5& 18.4M& 7.3G & 50.0& 53.4& 18.5M&6.8G \\
\hline
PosPool$^*$ (S) & 92.6& 1.2M& 0.1G& 61.3&  1.1M& 0.5G& 46.1& 47.2& 1.1M&0.5G\\
PosPool$^*$ &93.2 & 19.4M& 1.8G&66.7 & 18.4M& 7.3G& 50.6& 53.8& 18.5M& 6.8G\\
\Xhline{1.0pt}
\end{tabular}
\end{table}
\end{small}

Fig.~\ref{fig:arch} shows the residual architecture used in this paper. It consists 5 stages of different point resolution, with each stage stacked by several bottleneck residual blocks. Each bottleneck residual block is composed successively by a $1\times 1$ point-wise transformation layer, a local aggregation layer, and another $1\times 1$ point-wise transformation layer. At the block connecting two stages, a stridded local aggregation layer is applied where the local neighborhood is selected at a higher resolution and the output adopts a lower resolution. Batch normalization and ReLU layers are applied after each $1\times 1$ layer to facilitate training. For head networks, we use a 4-layer classifier and a U-Net style encoder-decoder~\cite{Ronneberger_2015} for classification and semantic segmentation, respectively.

In evaluation of a local aggregation operator, we use this operator to instantiate all local aggregation layers in the architecture. We also consider different model capacity by varying network depth (block repeating factor $N_r$), width ($C$) and bottleneck ratio ($\gamma$).

\vspace{0.3em} \noindent \textbf{Point Sampling and Neighborhoods.} To generate point sets for different resolution levels, we follow~\cite{Thomas_2018,thomas2019kpconv} to use a subsampling method with different grid sizes to generate point sets in different resolution stages. Specifically, the whole 3D space is divided by grids and one point is randomly sampled to represent a grid if multiple points appear in the grid. This method can alleviate the varying density problem~\cite{Thomas_2018,thomas2019kpconv}. Given a base grid size at the highest resolution of Res1, the grid size for different resolutions are multiplied by $2\times$ stage-by-stage. The base grid size for different datasets are detailed in Section~\ref{sec.exp}.

To generate a point neighborhood, we follow the ball radius method~\cite{qi2017pointnet++,Hermosilla_2018,liu2019rscnn}, which in general result in more balanced density than the location or feature kNN methods~\cite{wang2018paramconv,Atzmon_2018,wang2019dynamic}. The ball radius is set as $2.5\times$ of the base grid size.

\vspace{0.3em} \noindent \textbf{Datasets} We consider three datasets with varying scales of training data, task outputs (classification and semantic segmentation) and scenarios (CAD models and real scenes): ModelNet40~\cite{wu20153d}, S3DIS~\cite{2017arXiv170201105A} and PartNet~\cite{mo2019partnet}. More details about datasets are described in Section~\ref{sec.exp}.

\vspace{0.3em} \noindent \textbf{Performance of Two Baseline Operators}

For point cloud modeling, the architectures without local aggregation operators also perform well to some extent, e.g. PointNet~\cite{qi2017pointnet}. To investigate what local aggregation operators perform beyond, we present two baseline functions to replace the local aggregation operators described in Section~\ref{sec.review}:
\begin{small}
\begin{align}
&\mathbf{g}_i = \mathbf{f}_i, \label{eq.no_local} \\
&\mathbf{g}_i = R\left(\left\{\mathbf{f}_j | {j \in \mathcal{N}(i)}\right\}\right). \label{eq.max_avg_pool}
\end{align}
\end{small}The former is an identity function, without encoding neighborhood points. The latter is an AVG/MAX pool layer without regarding their relative positions.

Table~\ref{tab:baseline} shows the accuracy of these two baseline operators using the common architecture in Fig.~\ref{fig:arch} on three benchmarks. It can be seen that these baseline operators mostly perform marginally worse than the previous best performing methods on the three datasets. The baseline$^\dag$ operator using a MAX pooling layer even slightly outperforms the previous state-of-the-art with smaller computation FLOPs (47.4 mIoU, 6.7G FLOPs vs. 46.4 mIoU, 23.1G FLOPs).

In the following, we will revisit different designing components in the point-wise MLP based methods and the adaptive weight based methods using the common deep residual architecture in Fig.~\ref{fig:arch}. For the pseudo grid feature methods, we choose a representative operator, KPConv~\cite{thomas2019kpconv}, with depth-wise convolution kernel and its default grid settings ($M=15$) for comparison. There are not much hyper-settings for it and we will omit the detailed tuning.

\subsection{Performance Study on Point-wise MLP based Method}

\label{sec.benchmark_mlp}

We start the investigation of this type of methods from a representative method, PointNet++~\cite{qi2017pointnet++}. We first reproduce this method using its own specific overall architecture and with other implementation details the same as ours. Table~\ref{tab:pw_mlp} (denoted as PointNet++$^*$) shows our reproduction is fairly well, which achieves slightly better accuracy than that reported by the authors~\cite{qi2017pointnet++,mo2019partnet} on ModelNet40 and PartNet.

We re-investigate several design components for this type of methods using the deep architecture in Fig.~\ref{fig:arch}, including the number of fully connected (FC) layers in an MLP, the choice of input features and the reduction function. Table~\ref{tab:pw_mlp} shows the ablation study on these aspects, with architecture hyper-parameters as: block repeat factor $N_r = 1$, base width $C=144$ and bottleneck ratio $\gamma = 8$.

\begin{small}
\begin{table}[t]
    \centering
    \caption{Evaluating different settings of the point-wise MLP method. The option $\nabla$, $\triangle$, $\Box$ and $\Diamond$ denote input features using $\{\Delta \mathbf{p}_{ij}, \mathbf{f}_{j}\}$, $\{\mathbf{f}_{i}, \Delta \mathbf{f}_{ij}\}$, $\{\Delta \mathbf{p}_{ij}, \mathbf{f}_{i}, \Delta \mathbf{f}_{ij}\}$, and $\{\Delta \mathbf{p}_{ij}, \mathbf{f}_{i}, \mathbf{f}_{j}, \Delta \mathbf{f}_{ij}\}$, respectively. ``Sweet spot'' denotes balanced settings regarding both efficacy and efficiency. The accuracy on PartNet test set is not tested in ablations to avoid the tuning of test set}
    \label{tab:pw_mlp}
    \addtolength{\tabcolsep}{.1pt}
    \begin{tabular}{c|c|cccc|c|c|c|c|c}
\Xhline{1.0pt}
\multirow{2}{*}{method} & \multirow{2}{*}{$\gamma$} & \multicolumn{4}{c|}{input} & \multirow{2}{*}{\#FC} & \multirow{2}{*}{$R(\cdot)$} & \multirow{2}{*}{ModelNet40} & \multirow{2}{*}{S3DIS} & \multirow{2}{*}{\makecell{PartNet\\(val/test)}} \\
\cline{3-6}
& & $\nabla$ & $\triangle$ & $\Box$ & $\Diamond$ & & & & & \\
\hline
\makecell{PointNet++~\cite{qi2017pointnet++}} & - & \cmark & & & & 3 & MAX & 90.7 & - & -/42.5 \\
\makecell{PointNet++*} & - & \cmark & & & & 3 & MAX & 91.6 & 55.3 & 43.1/45.3 \\
\hline
\multirow{2}{*}{sweet spot}& 8 & & & \cmark & & 1 & MAX & 92.8 & 62.9 & 48.2/50.8 \\
 & 2 & & & \cmark & & 1 & MAX & 92.8 & 66.2 & 48.1/51.2 \\
\hline
\multirow{2}{*}{FC num} & 8 &  & & \cmark & & 2 & MAX & 92.5 & 59.5 & 47.9/- \\
& 8 &  & & \cmark & & 3 & MAX & 92.0 & 59.9 & 48.7/- \\
 \hline
 \multirow{3}{*}{input}& 8 & \cmark & &  & & 1 & MAX & 92.6 & 59.8 & 47.1/- \\
& 8 &  & \cmark & & & 1 & MAX & 92.5 & 61.4	& 47.6/- \\
& 8 &  & & &\cmark & 1 & MAX & 92.7	& 51.0	& 47.9/- \\
 \hline
 \multirow{2}{*}{reduction $R(\cdot)$}& 8 & & & \cmark & & 1 & AVG & 92.3&	55.1	&46.8/- \\
& 8 &  &  & \cmark& & 1 & SUM &92.2&	44.7&	46.7/- \\
\Xhline{1.0pt}
\end{tabular}
    \vspace{-2em}
\end{table}
\end{small}

We can draw the following conclusions:
\begin{itemize}
    \item \emph{Number of FC layers}. In literature of this method type, 3 layers are usually used by default to approximate complex functions. Surprisingly, in our experiments, \emph{using 1 FC layer without non-linearity significantly outperforms that using 2 or 3 FC layers on S3DIS}, and it is also competitive on ModelNet40 and PartNet. We hypothesize that the fitting ability by multiple FC layers applied on the concatenation of point feature and relative position may be partly realized by the point-wise transformation layers (the first and the last layers in a residual block) applied on point feature alone. Less FC layers also ease optimization. Using 1 FC layer is also favorable considering the efficiency issue: the computation can be significantly reduced when 1 FC layer is adopted, through computing sharing as explained below.
    \item \emph{Input Features}. The relative position and edge feature perform similarly on ModelNet40 and PartNet, and combining them has no additional gains. However, on S3DIS datasets, combining both significantly outperforms the variants using each alone.
    \item \emph{Reduction function}. MAX pooling performs the best, which is in accord with that in literature.
\end{itemize}

\vspace{0.3em} \noindent \textbf{An efficient implementation when 1 FC layer is used.} Denote the weight matrix of this only FC layer as $W = [W^1, W^2] \in \mathbb{R}^{d\times (d + 3)}$ where $W^1 \in \mathbb{R}^{d\times 3}$ and $W^2 \in \mathbb{R}^{d\times d}$. We have $G = W \cdot  \text{concat}(\Delta \mathbf{p}_{ij}, \mathbf{f}_j) = W^1\Delta \mathbf{p}_{ij}+ W^2\mathbf{f}_j$. Noting the computation of the second term $W^2\mathbf{f}_j$ can be shared when point $j$ appears in different neighborhoods, the computation complexity of this operator is significantly reduced from $(d + 3)ndK$ to $nd^2 +3ndK$.

\vspace{0.3em} \noindent \textbf{Sweet spots for point-wise MLP methods.} Regarding both the efficacy and efficiency, the sweet spot settings are applying 1 FC layer to an input combination of relative position and edge features. Table~\ref{tab:pw_mlp} also shows that using $\gamma = 2$ for this method can approach or surpass the state-of-the-art on all three datasets.

\subsection{Performance Study on Adaptive Weight based Method}

Table~\ref{tab:adaptive_weight} shows the ablations over several designing components within this method type, including the number of fully connected (FC) layers, choice of input features, the reduction function and whether to do weight normalization. We adopt architecture hyper-parameters as: block repeat factor $N_r = 1$, base width $C=144$, and bottleneck ratio $\gamma = 8$.

\begin{small}
\begin{table}[t]
    \centering
    \caption{Evaluating different settings of the adaptive weight based methods. $d\mathbf{p}*$ denotes the 9-dimensional position vector as in~\cite{liu2019rscnn}. ``Sweet spot'' denotes balanced settings regarding both efficacy and efficiency. The accuracy on PartNet test set is not tested for ablations to avoid tuning the test set.}
    \label{tab:adaptive_weight}
    \addtolength{\tabcolsep}{.1pt}
    \begin{tabular}{c|c|cccc|c|c|c|c|c|c}
\Xhline{1.0pt}
\multirow{2}{*}{method} & \multirow{2}{*}{$\gamma$} & \multicolumn{4}{c|}{input} & \multirow{2}{*}{\#FC} & \multirow{2}{*}{$R(\cdot)$} & \multirow{2}{*}{S.M.} & \multirow{2}{*}{ModelNet} & \multirow{2}{*}{S3DIS} & \multirow{2}{*}{\makecell{PartNet\\(val)}} \\
\cline{3-6}
& & $\{d\mathbf{p}\}$ & $\{d\mathbf{f}\}$ & $\{d\mathbf{p},d\mathbf{f}\}$ & $\{d\mathbf{p}^*\}$ & & & & & & \\
\hline
\makecell{PConv\cite{wang2018paramconv}} & - & \cmark & & & & 2 & SUM & & - & 58.3 & - \\
\makecell{FlexConv\cite{groh2018flex}} & - & \cmark & & & & 1 & SUM & & 90.2& 56.6 & - \\
\hline
sweet spot & 8 &  \cmark & & & & 1 & AVG & & 92.7&	62.6	&50.0\\
sweet spot* & 2 & \cmark & & & & 1 & AVG & & 93.0	&66.5	&50.1 \\
\hline
\multirow{2}{*}{FC num} & 8 & \cmark & &  & & 2 & AVG & & 92.6	&61.3	&49.9 \\
& 8 & \cmark & &  & & 3 & AVG & & 92.5	&58.5	&49.6 \\
 \hline
 \multirow{3}{*}{input}& 8 &  & \cmark &  & & 1 & AVG & & 85.3	&46.6&	46.9\\
& 8 &  &  & \cmark & & 1 & AVG & & 82.2	&55.7	&46.4 \\
& 8 &  & & &\cmark & 1 & AVG & & 92.1&	57.0	&49.1 \\
 \hline
 \multirow{2}{*}{reduction}& 8 &\cmark & &  & & 1 & SUM & & 92.6&	61.7	&49.1 \\
& 8 &\cmark  &  & & & 1 & MAX & & 92.4	&62.3	&49.7 \\
 \hline
 \multirow{1}{*}{SoftMax}& 8 & \cmark & &  & & 1 & AVG & \cmark &91.7	&55.9	&45.8 \\
\Xhline{1.0pt}
\end{tabular}
    \vspace{-2em}
\end{table}
\end{small}

We can draw the following conclusions:
\begin{itemize}
    \item \emph{Number of FC layers.} Using 1 FC layer performs noticeably better than that using 2 or 3 layers on S3DIS, and is comparable on ModelNet40 and PartNet.
    \item \emph{Input features.} Using relative positions alone performs best on all datasets. The accuracy slightly drops with additional position features~\cite{liu2019rscnn}. The edge features harm the performance, probably because it hinders the effective learning of adaptive weights from relative positions.
    \item \emph{Reduction function.} MAX and AVG functions perform slightly better than SUM function, probably because the MAX and AVG functions are more insensitive to varying neighbor size. We use AVG function by default.
    \item \emph{SoftMax normalization.} The accuracy significantly drops by SoftMax normalization, probably because the positive weights after normalization let kernels act as low-pass filters and may cause the over-smoothing problem~\cite{li2019can}.
\end{itemize}

\vspace{0.3em} \noindent \textbf{Sweet spots for adaptive weight based methods.} The best performance is achieved by applying 1 FC layer without SoftMax normalization on relative positions alone to compute the adaptive weights. This method also approaches or surpasses the state-of-the-art on all three datasets using a deep residual network.

\subsection{Discussions} Table~\ref{tab:baseline} indicates that the three local aggregation operator types with appropriate settings all achieve the state-of-the-art performance on three representative datasets using the same deep residual architectures. With $16\times$ less parameters and computations (marked by ``S''), they also perform competitive compared with the previous state-of-the-art. The sweet spots of different operators also favor a simplicity principle, that the relative position alone and 1 FC layer perform well in most scenarios.

While recent study in point cloud analysis mainly lies in inventing new local aggregation operators, the above results indicate that some of them may worth re-investigation under deeper and residual architectures. These results also stimulate a question: could a much simpler local aggregation operator achieve similar accuracy as the sophisticated ones? In the following section, we will try to answer this question by presenting an extremely simple local aggregation operator.

\section{PosPool: An Extremely Simple Local Aggregation Operator}

In this section, we present a new local aggregation operator, which is extremely simple with no learnable weights.

\begin{figure}[t]
\begin{center}
   \includegraphics[width=0.6\linewidth]{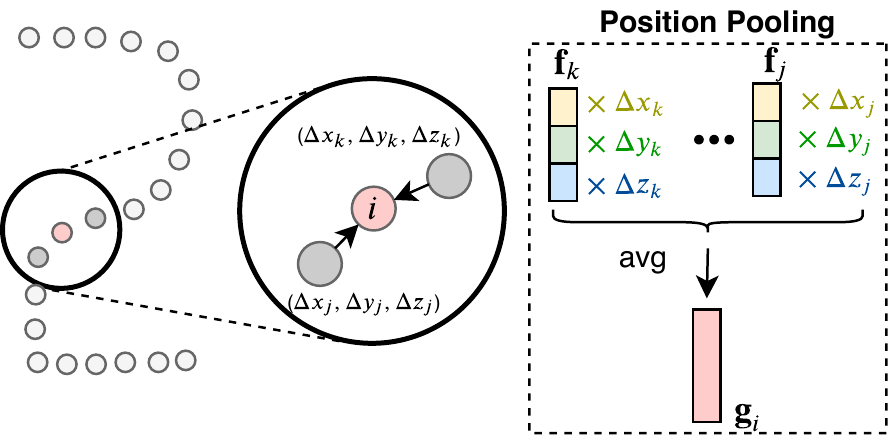}
\end{center}
   \vspace{-2em}
   \caption{Illustration of the proposed position pooling (PosPool) operator.}
   \vspace{-1em}
\label{fig:pos_pool}
\end{figure}

The new operator is illustrated in Fig.~\ref{fig:pos_pool}. For each neighboring point $j$, it combines the relative position $\Delta \mathbf{p}_{ij}$ and point feature $\mathbf{f}_{j}$ by element-wise multiplication. Considering the dimensional difference between the 3-dimensional $\Delta \mathbf{p}_{ij}$ and d-dimensional $\mathbf{f}_{j}$, the multiplication is applied group-wise that $\Delta \mathbf{p}_{ij}$'s three scalars $[\Delta x_{ij}, \Delta y_{ij}, \Delta z_{ij}]$ are multiplied to 1/3 channels of $\mathbf{f}_{j}$, respectively, as
\begin{equation}
\label{eq:pos_pool}
G(\Delta \mathbf{p}_{ij}, \mathbf{f}_j) = \text{Concat}\left[\Delta x_{ij} \mathbf{f}_j^0; \Delta y_{ij} \mathbf{f}_j^1; \Delta z_{ij} \mathbf{f}_j^2\right],
\end{equation}
where $\mathbf{f}_j^{0,1,2}$ are the 3 sub-vectors equally split from $\mathbf{f}_j$, as $\mathbf{f}_j = \left[\mathbf{f}_j^0; \mathbf{f}_j^1; \mathbf{f}_j^2\right]$.

The operator is named position pooling (PosPool), featured by its property of no learnable weight. It also reserves the \emph{permutation/translation invariance} property which is favorable for point cloud analysis.

\vspace{0.3em} \noindent \textbf{A Variant.} We also consider a variant of position pooling operator which is slightly more complex, but maintains the no learnable weight property. Instead of using 3-d relative coordinates, we embed the coordinates into a vector with the same dimension as point feature $\mathbf{f}_{ij}$ using cosine/sine functions, similar as in \cite{vaswani2017attention}. The embedding is concatenated from $d/6$ group of 6-dimensional vectors, with the $m^\text{th}$ 6-d vector representing the cosine/sine functions with a wave length of $1000^{6m/d}$ on relative locations $x,y,z$:
\begin{small}
\begin{align}
     \mathcal{E}^m(x, y, z)= & [sin(100x/1000^{6m/d}, cos(100x/1000^{6m/d}), \notag \\
    & sin(100y/1000^{6m/d}, cos(100y/1000^{6m/d}), \notag \\
    & sin(100z/1000^{6m/d}, cos(100z/1000^{6m/d})].
\end{align}
\end{small}

Then an element-wise multiplication operation is applied on the embedding $\mathcal{E}$ and the point feature $\mathbf{f}_{ij}$:
\begin{equation}
\label{eq:pos_pool_cosine_sine}
G(\Delta \mathbf{p}_{ij}, \mathbf{f}_j)=\mathcal{E} \odot \mathbf{f}_{ij}.
\end{equation}

The resulting operator also does not have any learnable weights, and is set as a variant of position pooling layer. We find this variant performs slightly better than the direct multiplication in Eq.~(\ref{eq:pos_pool}) in some scenarios. We will show more variants in Appendix A3.

\vspace{0.3em} \noindent \textbf{Complexity Analysis} The space complexity $\mathcal{O}(\text{space}) = 0$, as there are no learnable weights. The time complexity is also small $\mathcal{O}(\text{time}) = ndK$. Due to the no learnable weight nature, it may also potentially ease the hardware implementation, which does not require an adaption to different learnt weights.

\section{Experiments}
\label{sec.exp}

\subsection{Benchmark Settings}

In this section, we detailed the three benchmark datasets with varying scales of training data, task outputs (classification and semantic segmentation) and scenarios (CAD models and real scenes).

\begin{itemize}
    \item \emph{ModelNet40}~\cite{wu20153d} is a 3D classification benchmark. This dataset consists of 12,311 meshed CAD models from 40 classes. We follow the official data splitting scheme in~\cite{wu20153d} for training/testing. We adopt an input resolution of 5,000 and a base grid size of 2cm.
    \item \emph{S3DIS}~\cite{2017arXiv170201105A} is a real indoor scene segmentation dataset with 6 large scale indoor areas captured from 3 different buildings. 273 million points are annotated and classified into 13 classes. We follow~\cite{tchapmi2017segcloud} and use Area-5 as the test scene and all others for training. In both training and test, we segment small sub-clouds in spheres with radius of 2m. In training, the spheres are randomly selected in scenes. In test, we select spheres regularly in the point clouds. We adopt a base grid size of 4cm.
    \item \emph{PartNet}~\cite{mo2019partnet} is a more recent challenging benchmark for large-scale fine-grained part segmentation. This dataset consists of pre-sampled point clouds of $26,671$ 3D object models in $24$ object categories, with each object containing $18$ parts on average. This dataset is officially split into three parts: $70\%$ training, $10\%$ validation, and $20\%$ test sets. We train our model with official training dataset and then conduct the comparison study on the validation set on $17$ categories with fine-grained annotation. We also report the best accuracies of different methods on the test set. We use the 10,000 points provided with the datasets as input, and the base grid size is set as 2cm.
\end{itemize}

\begin{figure}[t]
\begin{center}
   \includegraphics[width=1.0\linewidth]{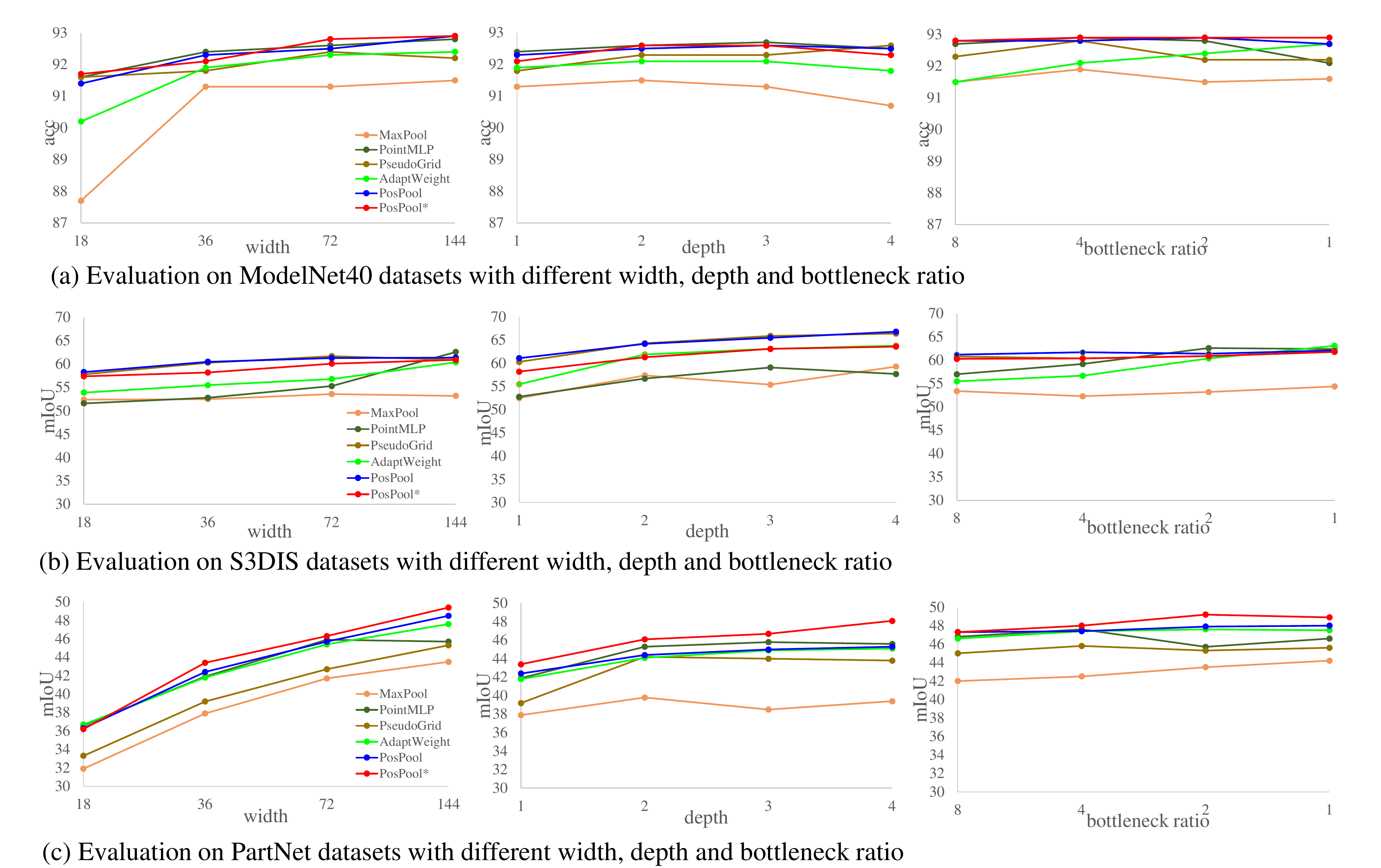}
\end{center}
   \vspace{-1em}
   \caption{Accuracy of different methods with varying width ($C$), depth ($N_r + 1$) and bottleneck ratio ($\gamma$) on three benchmark datasets.}
   \vspace{-1em}
\label{fig:vary_width_depth}
\end{figure}

\begin{figure}[t]
\begin{center}
   \includegraphics[width=0.95\linewidth]{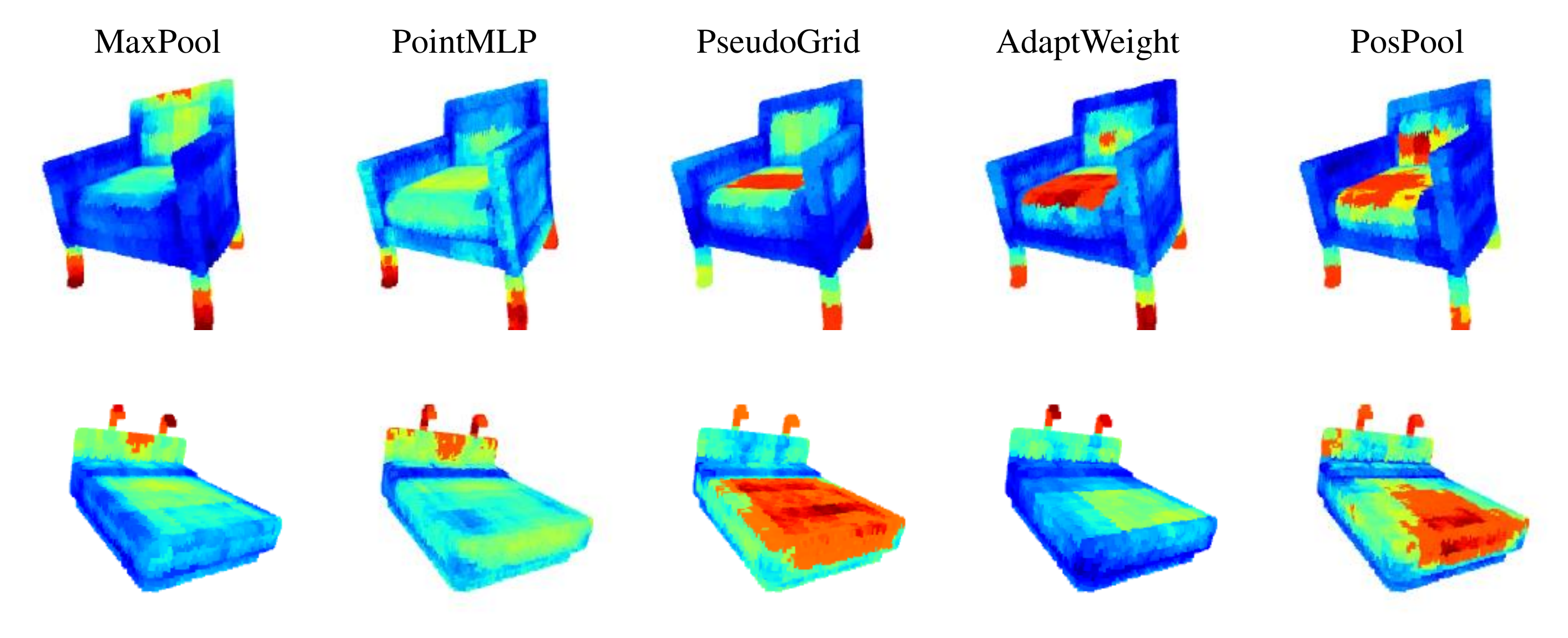}
\end{center}
   \vspace{-2em}
   \caption{Activation maps before the final prediction using different methods on PartNet validation shapes, indicating similar high energy area learnt by different methods}
   \vspace{-1em}
\label{fig:activation}
\end{figure}

The training/inference settings are detailed in Appendix A1. Note for PartNet datasets, while in~\cite{mo2019partnet} independent networks are trained for 17 different shapes, we adopt a shared backbone and independent 3 fully connected layers for part segmentation of different categories and train all the categories together, which significantly facilitate the evaluation on this dataset.  We note using the shared backbone network achieves similar accuracy than the methods training different shapes independently.

\subsection{Comparing Operators with Varying Architecture Capacity}

Fig.~\ref{fig:vary_width_depth} shows comparison of different local aggregation operators using architectures with different model capacity on three benchmarks, by varying the network width, depth and bottleneck ratio. Detailed experimental settings are presented in Appendix A2. It can be seen: the PosPool operators achieve top or close-to-top performances using varying network hyper-parameters on all datasets, showing its strong stability and adaptability. While the other more sophisticated operators may achieve similar accuracy with the PosPool layers on some datasets or settings, their performance are less stable across scenarios and model capacity. For example, the accuracy of the ``AdaptWeight'' method will drop significantly on S3DIS when the model capacity is reduced by either the width, depth or bottleneck ratio.
\vspace{-1em}

\section{Conclusion}
\vspace{-.5em}

This paper studies existing local aggregation operators in depth via a carefully designed common testbed that consists of a deep residual architecture and three representative benchmarks. Our investigation illustrates that with appropriate settings, all operators can achieve the state-of-the-art performance on three tasks. Motivated by this finding, we present a new extremely simple operator without learned weights, which performs as good as existing operators with sophisticated design. To understand what the networks with these operators are learnt from input, we visualize the norm of activation map before prediction by different methods (operators), suggesting that different operators tend to offer similar activations for a same input point cloud, as shown in Fig~\ref{fig:activation}. We hope our study and new design can encourage further rethinking and understanding on the role of local aggregation operators and shed new light to future network design.

%

\bibliographystyle{splncs04}
\bibliography{egbib}

\appendix

\renewcommand{\thesection}{A\arabic{section}}

\section{Training/Inference Settings}
In this section, we describe the training and inference settings of each dataset in detail.

\vspace{0.3em} \noindent \textbf{ModelNet40}
In training, we adopt the SGD optimizer, with initial learning rate of 0.002, which is decayed by $0.1^{1/200}$ every epoch. The momentum is 0.98 and weight decay is 0.001. The data is augmented with anisotropic random scaling (from 0.6 to 1.4), and gaussian noise of std = 0.002. We train networks for 600 epochs on 4 GPUs with 16 point clouds per GPU.

In inference, the model of the last epoch is used. We follow a common practice of voting scheme~\cite{qi2017pointnet++,wang2019dgcnn,liu2019rscnn,thomas2019kpconv}, which augment each shape 100 times using the same augmentation method in training, and the predicted logits (the values before SoftMax) of these 100 augmented shapes are averaged to produce the final proabalities.

\vspace{0.3em} \noindent \textbf{S3DIS}
Following~\cite{wang2018paramconv,li2018pointcnn,thomas2019kpconv}, we use 3 color channels as features.
In training, we adopt the SGD optimizer, with initial learning rate is 0.02,  which is decayed by $0.1^{1/200}$ every epoch. The momentum is 0.98 and weight decay is 0.001. The data is augmented with anisotropic random scaling (from 0.7 to 1.3),  gaussian noise of std = 0.001, random rotations around z-axis, random droping colors with 0.2 probability. The networks are trained for 600 epochs, using 4 GPUs and 8 point clouds per GPU.

In inference, the model of the last epoch is used. We divide each point cloud into regular overlaped spheres (totally 100). A point may appear in multiple spheres, and its logit (before SoftMax) is set as the average of this point's logits in different spheres.

\vspace{0.3em} \noindent \textbf{PartNet}
In training, we adopt the AdamW optimizer~\cite{loshchilov2018decoupled} with learning rate of 0.000625. The momentum is 0.98 and the weight decay is 0.001. The data is augmented with anisotropic random scaling (from 0.8 to 1.2), and gaussian noise of std = 0.001. The networks are trained for 300 epochs on 4 GPUs with 8 point clouds per GPU.

In inference, the model of the last epoch is used. We adopt the 10-augment voting scheme (using the same augmentation method as in training) to compute each point's probabilities.

\section{Detailed Experimental Settings for Section 6.2}

In Section 6.2 of the main paper, we evaluate different methods with varying architecture width, depth and bottleneck ratios. In this section, we provide detailed experimental settings as below.

For the experiments of varying architecture widths (see Fig. 3 left column of the main paper), we fix the depth and bottleneck ratio as $N_r+1 = 1$ and $\gamma=2$, respectively. For the experiments of varying architecture depths (see Fig. 3 middle column of the main paper), we fix the width and bottleneck ratio as $C = 36$ and $\gamma=2$, respectively. For the experiments of varying architecture bottleneck ratios (see Fig. 3 right column of the main paper), we fix the width and depth as $C = 144$ and $N_r+1 = 1$, respectively.

For different methods, the designing settings are as follows:
\begin{itemize}
    \item PointMLP. We use $\{\Delta \mathbf{p}_{ij}, \mathbf{f}_{i}, \Delta \mathbf{f}_{ij}\}$ as input features, MAX pooling as reduction function and 1 FC layer.
    \item PseudoGrid. We use SUM as reduction function and 15 grid points.
    \item AdaptWeight. We use $\{\Delta \mathbf{p}_{ij}\}$ as input features, AVG pooling as reduction function and 1 FC layer.
    \item PosPool. We use AVG pooling as reduction function and the computation follows Eq. 8 in our main paper.
    \item PosPool*.  We use AVG pooling as reduction function and the computation follows Eq. 9 in our main paper.
\end{itemize}

\section{More Variants of PosPool}

In this section, we present more variants for PosPool, which all have no learnable weights. We first present a general formulation of these variants:
\begin{small}
	\begin{align}
	G(\Delta \mathbf{p}_{ij}, \mathbf{f}_j) = \text{Concat}[e^0  \mathbf{f}_j^0; ...; e^{g-1} \mathbf{f}_j^{g-1}],
	\end{align}
\end{small}where $\{e^0, ..., e^{g-1}\}$ are $g$ scalar encoding functions w.r.t. the relative position; $\mathbf{f}_j = \left[\mathbf{f}_j^0; \mathbf{f}_j^0; ...;\mathbf{f}_j^{g-1}\right]$ are an equal-sized partition of vector $\mathbf{f}_j$. In the following, we will present 7 variants by using different encoding functions $\{e^0, ..., e^{g-1}\}$.

\vspace{0.3em} \noindent \textbf{Second Order} Instead of directly using the 3-dimensional $\Delta \mathbf{p}_{ij}$ as in the standard formulation of Eq.~8 in the main paper, the second-order variant considers 6 additional encoding scalars by squares and pairwise multiplications of relative coordinates, as
\begin{small}
	\begin{align}
	& e^0 = \Delta x_{ij}, e^1 = \Delta y_{ij}, e^2 = \Delta z_{ij}, \notag   \\
	&e^3 = \Delta x_{ij}^2, e^4 = \Delta y_{ij}^2, e^5 = \Delta z_{ij}^2, \notag  \\
	&e^6 = \Delta x_{ij} \Delta y_{ij}, e^7 = \Delta x_{ij} \Delta z_{ij}, e^8 = \Delta y_{ij} \Delta z_{ij}.
	\end{align}
\end{small}

\vspace{0.3em} \noindent \textbf{Third Order} The third order variant uses additional third-order multiplications as encoding functions:
\begin{small}
	\begin{align}
	& e^0 = \Delta x_{ij}, e^1 = \Delta y_{ij}, e^2 = \Delta z_{ij},
	e^3 = \Delta x_{ij}^2, e^4 = \Delta y_{ij}^2, e^5 = \Delta z_{ij}^2, \notag  \\
	&e^6 = \Delta x_{ij} \Delta y_{ij}, e^7 = \Delta x_{ij} \Delta z_{ij}, e^8 = \Delta y_{ij} \Delta z_{ij}, \notag \\
	&e^9 = \Delta x_{ij}\Delta y_{ij}^{2}, e^{10} = \Delta x_{ij}\Delta z_{ij}^{2}, e^{11} = \Delta y_{ij}\Delta z_{ij}^{2}, \notag \\
	&e^{12} = \Delta x_{ij}^2 \Delta y_{ij}, e^{13} = \Delta x_{ij}^2 \Delta z_{ij}, e^{14} = \Delta y_{ij}^2 \Delta z_{ij}, \notag \\
	&e^{15} = \Delta x_{ij}^{3}, e^{16} = \Delta y_{ij}^{3}, e^{17} = \Delta z_{ij}^{3}.
	\end{align}
\end{small}Note we omit the encoding function $\Delta x_{ij} \Delta y_{ij}\Delta z_{ij}$ to ensure $g=18$ such that $\mathbf{f}_j$'s channel number $C$ is divisible by $g$. The third order encoding functions are similar as the Taylor functions in~\cite{xu2018spidercnn}.

\vspace{0.3em} \noindent \textbf{Angle and Distance} In this variant, we decouple the relative position into distance $d_{ij}=\sqrt{\Delta x^2_{ij} + \Delta y^2_{ij} + \Delta z^2_{ij}}$ and angle $\left \{\frac{\Delta x_{ij}}{d_{ij}}, \frac{\Delta y_{ij}}{d_{ij}}, \frac{\Delta z_{ij}}{d_{ij}} \right\}$. The encoding functions are:
\begin{small}
	\begin{align}
	& e^0 = d_{ij}, e^1 = \frac{\Delta x_{ij}}{d_{ij}}, e^2 = \frac{\Delta y_{ij}}{d_{ij}}, e^3 = \frac{\Delta z_{ij}}{d_{ij}}.
	\end{align}
\end{small}

\vspace{0.3em} \noindent \textbf{Angle and Gaussian Inversed Distance} The above variants encourage the distant points to have larger amplitudes of encoding scalars. Here we present a variant which encourages close points to have larger amplitudes of encoding scalars, by inverse the distance by a Gaussian function:
\begin{small}
	\begin{align}
	& e^0 = \tilde{d}_{ij}=\exp\left(-{d_{ij}^{2}}\right), e^1 = \frac{\Delta x_{ij}}{d_{ij}}, e^2 = \frac{\Delta y_{ij}}{d_{ij}}, e^3 = \frac{\Delta z_{ij}}{d_{ij}}.
	\end{align}
\end{small}

\vspace{0.3em} \noindent \textbf{Angle or Distance Alone} We also consider variants which use angle or distance functions alone:
\begin{small}
	\begin{align}
	& e^0 = \frac{\Delta x_{ij}}{d_{ij}}, e^1 = \frac{\Delta y_{ij}}{d_{ij}}, e^2 = \frac{\Delta z_{ij}}{d_{ij}}. \\
	& e^0 =  d_{ij}. \\
	& e^0 = \tilde{d}_{ij}.
	\end{align}
\end{small}

\vspace{0.3em} \noindent \textbf{Results.} Table~\ref{tab:pos_pool} shows the comparison of different variants using three benchmarks. For PartNet datasets, we report the part category mean IoU on the validation set. While PosPool adopts AVG as the default reduction function, we also report the results when using other reduction function (SUM, MAX). It can be seen: 1) all variants containing full configurations of relative positions perform similarly well. They perform significantly better than the variants using angle or distance alone. 2) Whether more distant points have larger or smaller encoding amplitudes than closer points is insignificant. 3) Using AVG as the reduction function performs comparably well than those using SUM, and slightly better than those using MAX.

\begin{small}
	\begin{table}[t]
		\centering
		\caption{Evaluation of different PosPool variants on three benchmarks. For PartNet datasets, the part category mean IoU on validation set are reported}
		\label{tab:pos_pool}
		\addtolength{\tabcolsep}{.1pt}
		\begin{tabular}{c|c|c|c|c|c|c|c}
			\Xhline{1.0pt}
			method&$\gamma$&$C$&$N_r+1$&$R(\cdot)$&ModelNet40&S3DIS&PartNet\\
			\hline
			\multirow{5}{*}{$\Delta x_{ij}, \Delta y_{ij}, \Delta z_{ij}$} & 8 &144&2& AVG & 93.0 & 64.2 &  48.5\\
			& 4 &144&2&AVG & 92.8 & 65.1 &  50.0\\
			& 2 &144&2&AVG & 93.1 & 66.6 &49.8 \\
			& 8 &144&2& SUM & 92.8 & 64.6 &  48.5\\
			& 8 &144&2& MAX & 92.6 & 61.1 & 48.4\\
			\hline
			\multirow{5}{*}{$\mathcal{E}^m$}  & 8 &144&2& AVG & 92.7 & 62.2 & 49.3 \\
			& 4 &144&2& AVG & 93.1 & 63.6 & 50.9 \\
			& 2 &144&2& AVG & 93.2 & 64.9 & 50.8 \\
			& 8 &144&2& SUM & 92.8 & 62.9& 49.1 \\
			& 8 &144&2& MAX & 92.4 & 62.8 & 48.9 \\
			\hline
			\multirow{5}{*}{Second Order} & 8 &144&2& AVG & 93.0 & 63.4 & 49.9 \\
			& 4 &144&2& AVG & 93.1 & 64.0 & 49.9 \\
			& 2 &144&2& AVG & 92.9 & 65.7 & 50.9 \\
			& 8 &144&2& SUM & 92.9 & 64.0 & 49.9 \\
			& 8 &144&2& MAX & 92.7 & 63.3 & 48.1 \\
			\hline
			\multirow{5}{*}{Third Order} & 8 &144&2& AVG & 93.2 & 63.6 & 49.6\\
			& 4 &144&2& AVG & 93.3 & 64.5 & 50.0 \\
			& 2 &144&2& AVG &  93.4& 64.7 & 51.8 \\
			& 8 &144&2& SUM & 92.7 &  64.8&  47.7\\
			& 8 &144&2& MAX & 92.3 &  62.2&  49.0\\
			\hline
			\multirow{5}{*}{$d_{ij}, \frac{\Delta x_{ij}}{d_{ij}}, \frac{\Delta y_{ij}}{d_{ij}}, \frac{\Delta z_{ij}}{d_{ij}}$} & 8 &144&2& AVG & 92.8 & 63.5 & 49.0 \\
			& 4 &144&2& AVG & 93.2 & 65.3 & 48.3 \\
			& 2 &144&2& AVG & 92.9 & 65.6 & 49.8 \\
			& 8 &144&2& SUM & 92.9 & 64.5 & 49.0 \\
			& 8 &144&2& MAX & 92.7 & 62.3 & 48.4 \\
			\hline
			\multirow{5}{*}{$\tilde{d}_{ij}, \frac{\Delta x_{ij}}{d_{ij}}, \frac{\Delta y_{ij}}{d_{ij}}, \frac{\Delta z_{ij}}{d_{ij}}$} & 8 &144&2& AVG & 93.0 & 64.2 & 48.6 \\
			& 4 &144&2& AVG & 93.2 & 64.1 & 49.1 \\
			& 2 &144&2& AVG & 93.0 & 64.8 & 49.3 \\
			& 8 &144&2& SUM & 93.0 & 64.3 & 48.2 \\
			& 8 &144&2& MAX & 92.9 & 62.3 & 49.1 \\
			\hline
			\multirow{5}{*}{$\frac{\Delta x_{ij}}{d_{ij}}, \frac{\Delta y_{ij}}{d_{ij}}, \frac{\Delta z_{ij}}{d_{ij}}$} & 8 &144&2& AVG &92.1 &62.1  &46.6 \\
			& 4 &144&2& AVG &92.1 & 61.8 &  46.5\\
			& 2 &144&2& AVG &  92.2&  62.6& 47.6 \\
			& 8 &144&2& SUM & 91.9 &  60.9& 46.4 \\
			& 8 &144&2& MAX & 92.0 &  61.2&  45.8\\
			\hline
			\multirow{5}{*}{$d_{ij}$} & 8 &144&2 & AVG & 90.9& 53.3&  43.4\\
			& 4 &144&2& AVG &  91.2& 53.1&  43.8\\
			& 2 &144&2& AVG &  90.9&  53.4&  44.6\\
			& 8 &144&2& SUM &  90.7&  55.4&43.6  \\
			& 8 &144&2& MAX &  91.0& 56.2 &44.2  \\
			\hline
			\multirow{5}{*}{$\tilde{d}_{ij}$} & 8 &144&2& AVG & 90.6&53.7 &  43.7\\
			& 4 &144&2& AVG &  90.5& 53.0&  42.1\\
			& 2 &144&2& AVG &  90.7&  53.4 &  43.9\\
			& 8 &144&2& SUM &  91.1&  55.3& 45.4 \\
			& 8 &144&2& MAX & 91.7 &  55.5&43.4  \\
			\Xhline{1.0pt}
		\end{tabular}
		\vspace{-2em}
	\end{table}
\end{small}

\section{The Robustness of Different Operators with Missing/Noisy Points}
\begin{figure}[t]
	\begin{center}
		\includegraphics[width=1.0\linewidth]{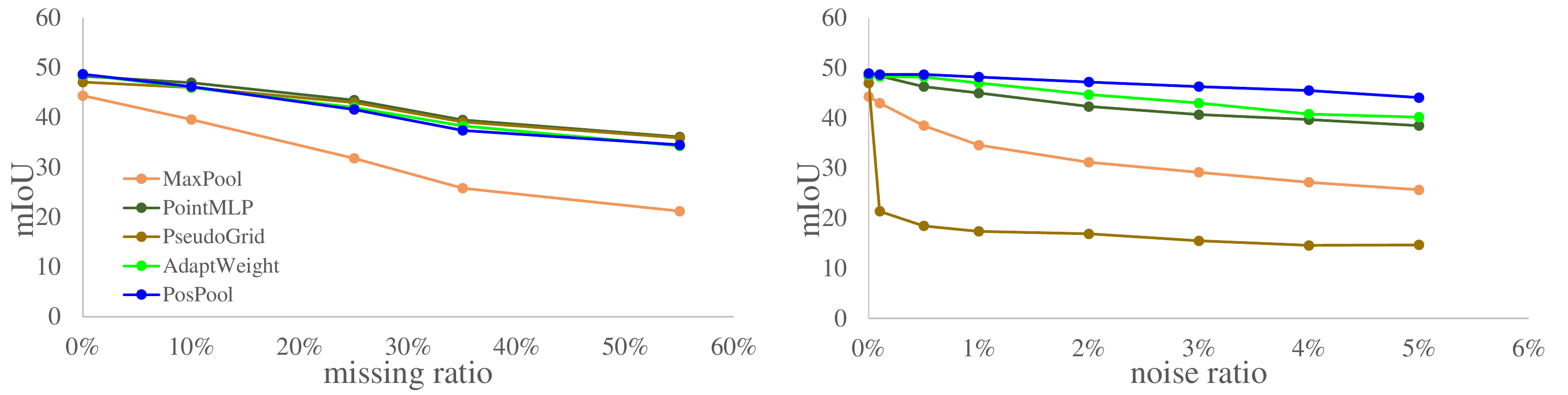}
	\end{center}
	\vspace{-2em}
	\caption{The robustness test of different approaches when there are less points(left) or outlier points(right).}
	\vspace{-1em}
	\label{fig:robust}
\end{figure}

Fig~\ref{fig:robust} show the accuracy curves of AdaptWeight, PseudoGrid, MaxPool, PosPool, PointMLP for inputs with different ratios of noise or missing points. All the experiments are executed on the PartNet benchmark. The model for each curve is trained on the clean data of PartNet. Only the testing data at the inference stage includes noise and missing points.

\begin{figure}[t]
	\begin{center}
		\includegraphics[width=0.8\linewidth]{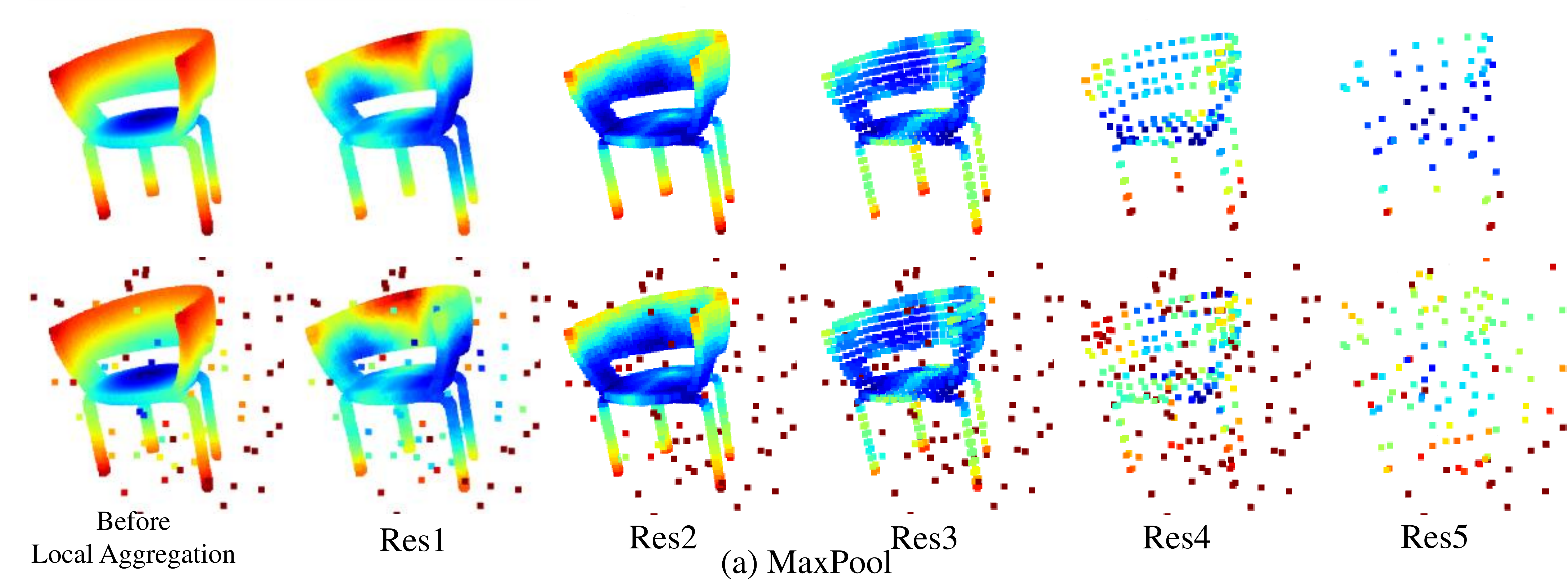} \\
		\includegraphics[width=0.8\linewidth]{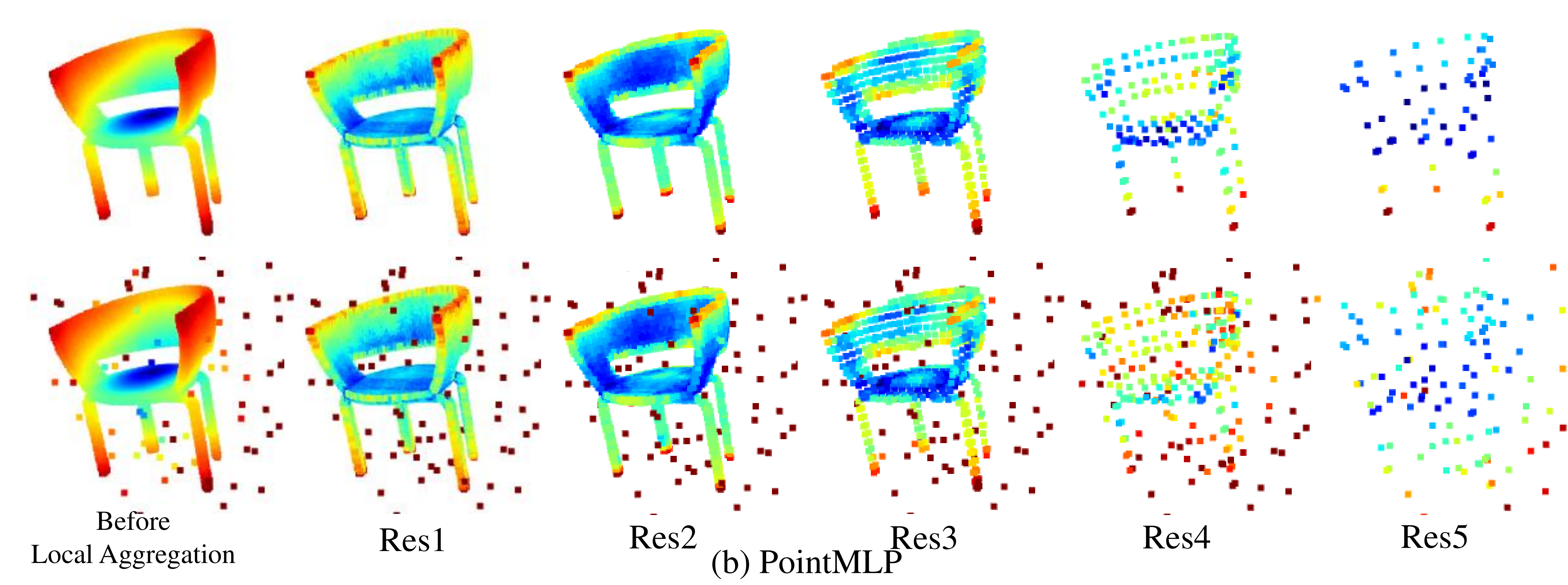} \\
		\includegraphics[width=0.8\linewidth]{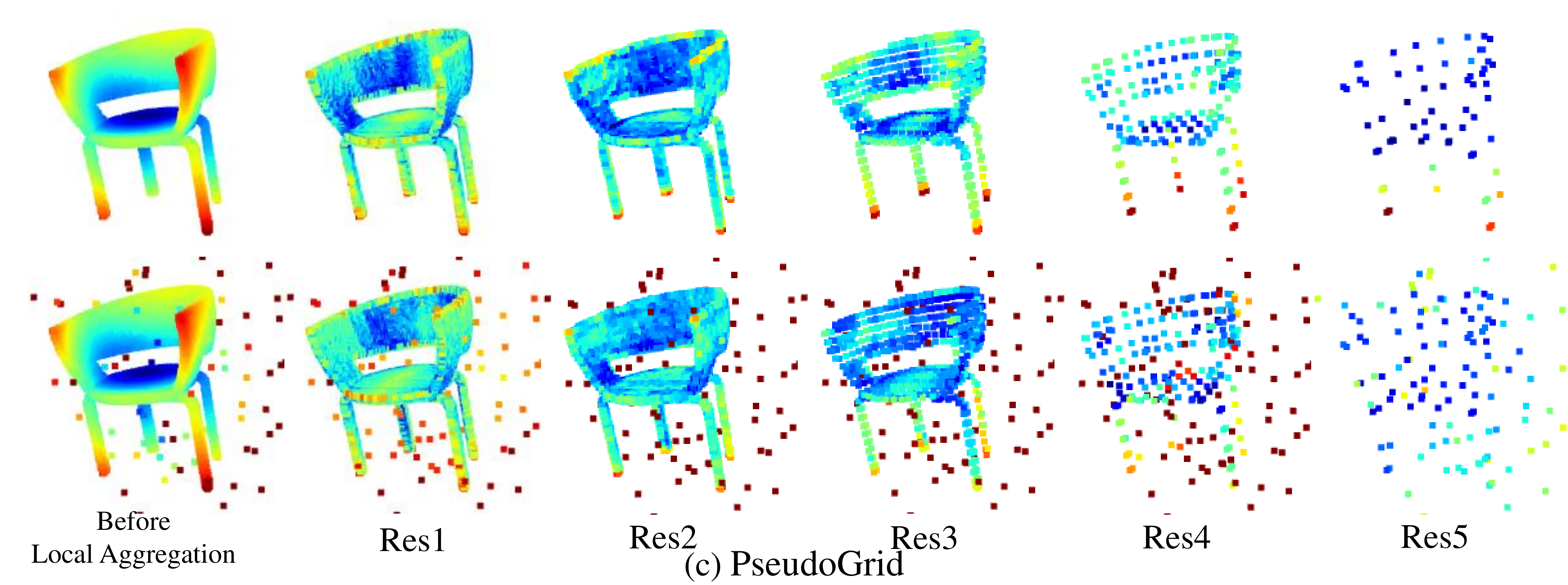} \\
		\includegraphics[width=0.8\linewidth]{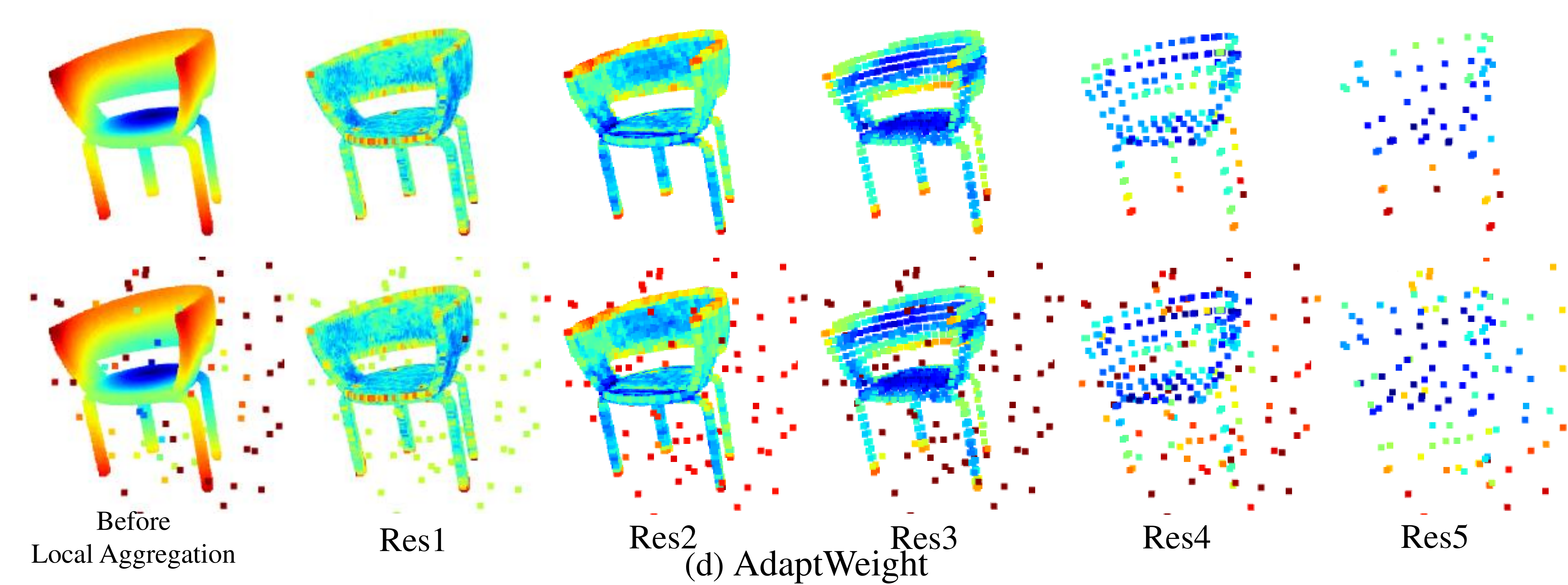} \\
		\includegraphics[width=0.8\linewidth]{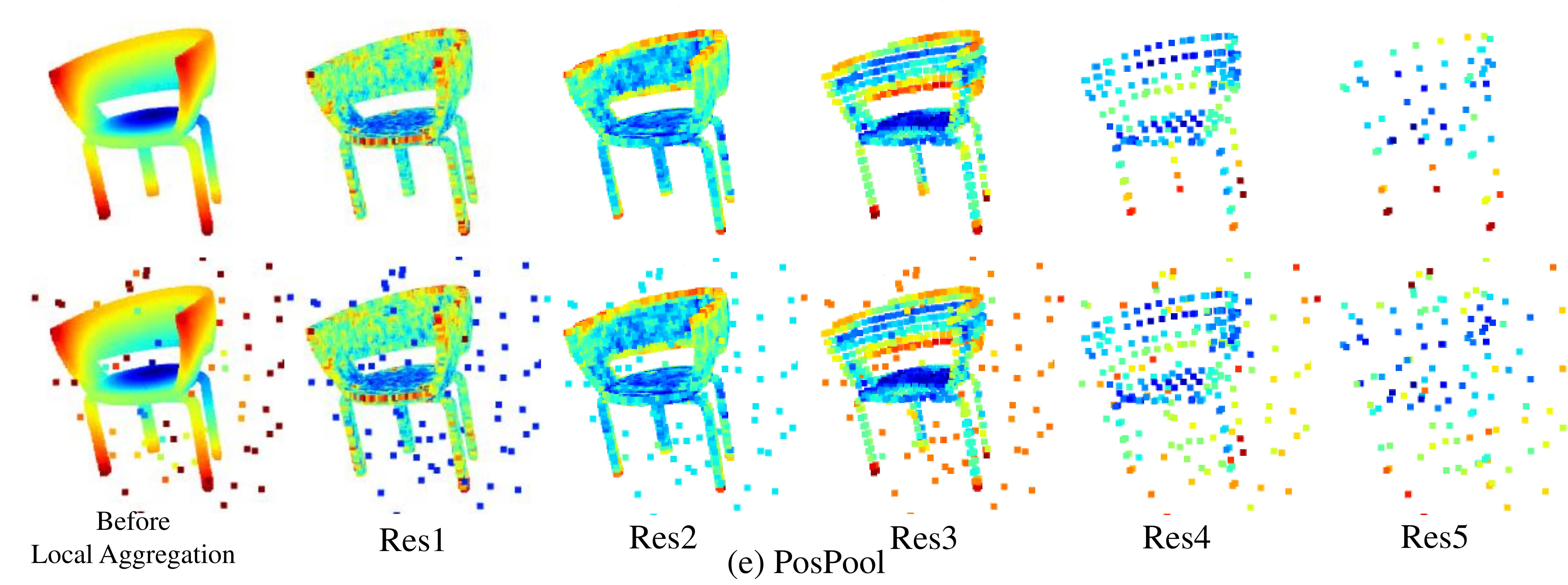} \\
	\end{center}
	\vspace{-2em}
	\caption{The activation maps of different methods by using clean data(top) or noisy data(bottom), respectively}
	\vspace{-1em}
	\label{fig:noise_act}
\end{figure}

As shown in Fig~\ref{fig:robust}(left), different local aggregation operators (AdaptWeight, PseudoGrid, PointMLP, PosPool) perform similarly in robustness with varying missing point ratios, all significantly better than the MaxPool baseline. With varying noise ratios, the proposed PosPool operator performs best, slightly better than AdaptWeight and PointMLP, and significantly better than PseudoGrid and the MaxPool baseline. Fig.~\ref{fig:noise_act} show the activation maps of the last layer in each stage by using clean data (top row) and noisy data (bottom row, ratio 1\%), respectively. While the noisy point features significantly contaminate the activations of some other regular point features in the MaxPool and PseudoGrid methods, the activations of clean points in other methods are less affected by these noisy points.

\section{More Detailed Results on PartNet}
We first report the part-category mIoU for each category on PartNet.  From Table~\ref{tab:partnetval} and Table~\ref{tab:partnettest} we can see that all operators show similar results on each category, which further validates our findings. Table~\ref{tab:partnetsplit} shows the number of training, validation, test samples.

\begin{figure}[t]
	\begin{center}
		\includegraphics[width=1.0\linewidth]{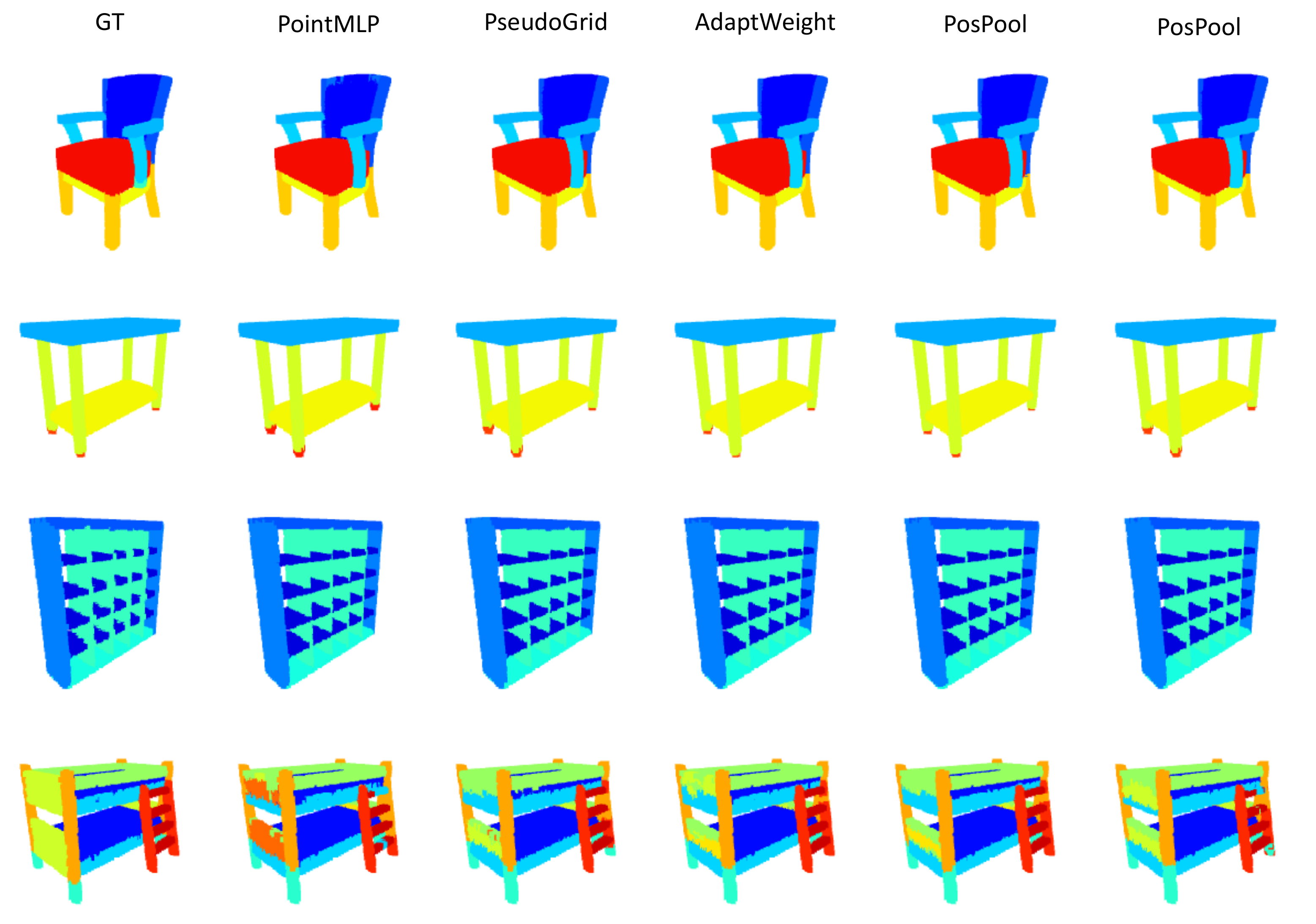}
	\end{center}
	\vspace{-2em}
	\caption{Visualization of part segmentation results by different methods on the PartNet dataset.}
	\vspace{-1em}
	\label{fig:quan_vis}
\end{figure}

We then show some qualitative results in Fig~\ref{fig:quan_vis}. All the representative methods perform similarly well on most shapes.

\begin{table*}[htb]
	\tiny
	\centering
	\renewcommand{\tabcolsep}{1pt}
	\begin{tabular}[width=\linewidth]{|l|c|ccccccccccccccccc|}
		\Xhline{2\arrayrulewidth}
		& \textbf{Avg}  & \textbf{Bed} & \textbf{Bott} & \textbf{Chair} & \textbf{Clock} & \textbf{Dish} & \textbf{Disp} & \textbf{Door} & \textbf{Ear} & \textbf{Fauc}  & \textbf{Knife} & \textbf{Lamp} & \textbf{Micro} & \textbf{Frid} & \textbf{Stora} & \textbf{Table} & \textbf{Trash} & \textbf{Vase} \\
		\Xhline{2\arrayrulewidth}
		\textbf{PW}  & 48.1 & 41.5 & 29.3 & 48.2 & 46.2 & 61.3 & 87.0 & 38.1 & 56.5 & 52.3 & 26.7 & 30.1 & 47.9 & 46.5 & 48.1 & 44.3 & 51.3 & 62.4  \\
		\textbf{PG}  & 50.8 &46.8 & 30.0 & 50.5 & 46.1 & 60.4 & 88.5 & 45.4 & 61.4 & 54.2 & 30.5 & 31.1 & 61.8 & 47.9 & 50.6 & 47.2 & 56.8 & 54.6  \\
		\textbf{AW}  &50.1 &  45.4 & 25.0 & 49.0 & 46.8 & 63.0 & 87.3 & 38.1 & 63.6 & 54.4 & 38.0 & 30.1 & 57.5 & 49.6 & 48.2 & 45.9 & 54.7 & 55.7  \\
		\textbf{PP} & 50.0 & 46.6 & 28.5 & 49.2 & 47.2 & 60.7 & 86.7 & 39.8 & 55.2 & 54.0 & 41.5 & 31.5 & 58.1 & 48.3 & 48.4 & 45.6 & 57.1 &  51.4  \\
		\textbf{PP$^*$} &50.6 & 47.5 & 29.7 & 49.1 & 47.2 & 65.8 & 88.0 & 46.8 & 58.9 & 54.6 & 31.5 & 28.1 & 60.7 & 47.3 & 50.9 & 45.0 & 54.6 & 55.0  \\
		\hline
	\end{tabular}
	\caption{\textbf{part-category mIoU\%} on PartNet validation sets.  \textbf{PW}, \textbf{PG}, \textbf{AW}, \textbf{PP}, \textbf{PP$^*$} refer to Pseudo Grid, Adapt Weights, PosPool, PosPool$^*$ respectively. }
	\label{tab:partnetval}
	\vspace{-3mm}
\end{table*}

\begin{table*}[htb]
	\tiny
	\centering
	\renewcommand{\tabcolsep}{1pt}
	\begin{tabular}[width=\linewidth]{|l|c|ccccccccccccccccc|}
		\Xhline{2\arrayrulewidth}
		& \textbf{Avg}  & \textbf{Bed} & \textbf{Bott} & \textbf{Chair} & \textbf{Clock} & \textbf{Dish} & \textbf{Disp} & \textbf{Door} & \textbf{Ear} & \textbf{Fauc}  & \textbf{Knife} & \textbf{Lamp} & \textbf{Micro} & \textbf{Frid} & \textbf{Stora} & \textbf{Table} & \textbf{Trash} & \textbf{Vase} \\
		\Xhline{2\arrayrulewidth}
		\textbf{PW} &51.2 & 44.5 & 52.6 & 46.0 & 38.4 & 68.2 & 82.5 & 46.9 & 47.1 & 58.7 & 43.8 & 26.4 & 59.2 & 48.7 & 52.5 & 41.3 & 55.4 & 57.3  \\
		\textbf{PG} & 53.0 & 47.5 & 50.9 & 49.2 & 44.8 & 67.0 & 84.2 & 49.1 & 49.9 & 62.7 & 38.3 & 27.0 & 59.4 & 54.3 & 54.1 & 44.5 & 57.4 & 60.7 \\
		\textbf{AW} &53.5 &  46.1 & 47.9 & 47.2 & 42.7 & 64.4 & 83.7 & 55.6 & 49.5 & 61.7 & 49.5 & 27.4 & 59.3 & 57.7 & 53.5 & 45.1 & 57.5 & 60.9  \\
		\textbf{PP} & 53.4 & 45.8 & 46.5 & 48.3 & 40.2 & 66.1 & 84.2 & 49.4 & 51.6 & 63.5 & 48.1 & 27.9 & 62.3 & 56.1 & 53.3 & 43.4 & 58.7 & 62.4  \\
		\textbf{PP$^*$} & 53.8 & 49.5 & 49.4 & 48.3 & 49.0 & 65.6 & 84.2 & 56.8 & 53.8 & 62.4 & 39.3 &  24.7 & 61.3 & 55.5 & 54.6 & 44.8 & 56.9 & 58.2  \\
		\hline
	\end{tabular}
	\caption{\textbf{part-category mIoU\%} on PartNet test sets. \textbf{PW}, \textbf{PG}, \textbf{AW}, \textbf{PP}, \textbf{PP$^*$} refer to Pseudo Grid, Adapt Weights, PosPool, PosPool$^*$ respectively.}
	\label{tab:partnettest}
	\vspace{-3mm}
\end{table*}

\begin{table*}[htb]
	\tiny
	\centering
	\renewcommand{\tabcolsep}{1pt}
	\begin{tabular}[width=\linewidth]{|l|ccccccccccccccccc|}
		\Xhline{2\arrayrulewidth}
		\textbf{Split}  & \textbf{Bed} & \textbf{Bott} & \textbf{Chair} & \textbf{Clock} & \textbf{Dish} & \textbf{Disp} & \textbf{Door} & \textbf{Ear} & \textbf{Fauc}  & \textbf{Knife} & \textbf{Lamp} & \textbf{Micro} & \textbf{Frid} & \textbf{Stora} & \textbf{Table} & \textbf{Trash} & \textbf{Vase} \\
		\Xhline{2\arrayrulewidth}
		\textbf{train} & 133 & 315 & 4489 & 406 & 111 & 633 & 149 & 147 & 435 & 221 & 1554 & 133 & 136 & 1588 & 5707 & 221 & 741\\
		\textbf{val} & 24 & 37 & 617 & 50 & 19 & 104 & 25 & 28 & 81 & 29 & 234 & 12 & 20 & 230 & 843 & 37 & 102\\
		\textbf{test} & 37 & 84 & 1217 & 98 & 51 & 191 & 51 & 53 & 132 & 77 & 419 & 39 & 31 & 451 & 1668 & 63 & 233\\
		\hline
	\end{tabular}
	\caption{The number of training, validation and test  samples.}
	\label{tab:partnetsplit}
	\vspace{-3mm}
\end{table*}

\section{Detailed Space and Time complexity Analysis}
In this section, we provide detailed analysis for the space and time complexity of different aggregators presented in Section 3.

\subsection{Point-wise MLP based Methods} 
\begin{figure}[t]
	\begin{center}
		\includegraphics[width=1.0\linewidth]{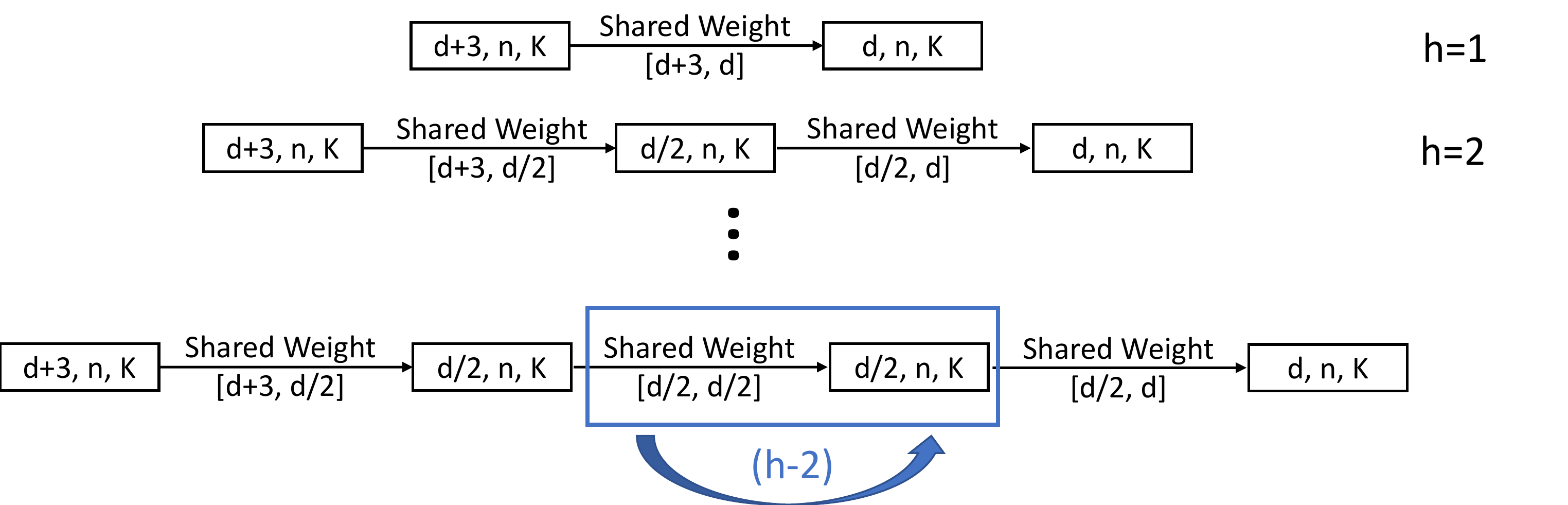}
	\end{center}
	\vspace{-2em}
	\caption{The detailed architecture for Point-wise MLP based operators.}
	\vspace{-1em}
	\label{fig:prof_pointwise}
\end{figure}

The detailed architecture is shown in Fig~\ref{fig:prof_pointwise}. For $h=1$, a shared FC is applied on each point with $K$ neighborhoods, and the time cost is $(d + 3)dnK$ and the space cost (parameter number) is $(d+3)d$. For $h \geq 2$, the time cost is
\begin{small}
	\begin{align}
	(d+3)(d/2)nK + (h-2) \cdot (d/2)(d/2)nK &+ (d/2)dnK=\notag \\
	& ((2d + 3) +(h-2)d/2)\cdot d/2 \cdot n K,
	\end{align}
\end{small}
and the space cost is
\begin{small}
	\begin{align}
	(d+3)(d/2) + (h-2) \cdot (d/2)(d/2) &+ (d/2)d = ((2d + 3) +(h-2)d/2)\cdot d/2.
	\end{align}
\end{small}

\subsection{Pseudo Grid Feature based Methods}
Our default settings adopt depth-wise convolution. In depth-wise convolution, a $d$-dim learnt weight vector is associated to each grid point. Hence, the space cost (parameter number) is $d\cdot M$ and the time cost is $ndKM$.

\subsection{Adaptive Weight based Methods}
Adaptive Weight based Methods involve two computation steps. Firstly, a shared MLP is used to compute the aggregation weights for each neighboring point. This step has time cost of
\begin{small}
	\begin{align}
	3\cdot(d/2)nK + (h-2) \cdot (d/2)(d/2)nK &+ (d/2)dnK=\notag \\
	& (3 + d +(h-2)d/2)\cdot d/2 \cdot n K,
	\end{align}
\end{small}
and space cost of
\begin{small}
	\begin{align}
	3\cdot (d/2) + (h-2) \cdot (d/2)(d/2) &+ (d/2)d = (3 + d +(h-2)d/2)\cdot d/2.
	\end{align}
\end{small}
Secondly, depth-wise aggregation is conducted, where the time cost is $dnK$ and space cost is $0$.

The total time cost is $(3 + d +(h-2)d/2)\cdot d/2 \cdot n K + dnK = ((h-2)d/2 + d + 5) \cdot d/2 \cdot nK$ and the total space cost is $ ((h-2)d/2 + d + 3) \cdot d/2$.

\end{document}